%% file: main.tex
\documentclass[letterpaper, 10 pt, conference]{ieeeconf}
\IEEEoverridecommandlockouts
\usepackage{graphics}
\usepackage{epsfig}
\usepackage{mathptmx}
\usepackage{times}
\usepackage{amsmath}
\usepackage{amssymb}
\usepackage{multicol}
\usepackage[bookmarks=true]{hyperref}
\usepackage{graphicx} 
\usepackage{float}
\usepackage{wrapfig}
\usepackage{lipsum}
\usepackage{float}
\restylefloat{figure}
\graphicspath{{./im/}}
\usepackage[font=small]{caption}
\usepackage[font=small]{subcaption}
\usepackage[draft]{pgf}
\usepackage{mathtools}
\usepackage{amsmath}
\usepackage{hyperref}
\usepackage{url}
\usepackage{flushend}
\usepackage{color}

\newcommand{\condition}[1]{\textbf{#1}}  
  
\title{\LARGE \bf
Experimental Assessment of Human-Robot Teaming 
for Multi-Step Remote Manipulation 
with Expert Operators}

\author{Claudia P\'{e}rez-D'Arpino, Rebecca P. Khurshid, Julie A. Shah
\thanks{Authors are with Computer Science and Artificial Intelligence Laboratory (CSAIL), Massachusetts Institute of Technology
{\tt\small \{cdarpino,rkhursh,julie\_a\_shah\}@csail.mit.edu}
}
}

\usepackage{fancyhdr}
\begin{document}
\belowdisplayskip=4pt plus 3pt minus 4pt
\belowdisplayshortskip=2pt plus 3pt minus 4pt

\maketitle
\thispagestyle{fancy}
\pagestyle{fancy}

\begin{abstract}
Remote robot manipulation with human control enables applications where safety and environmental constraints are adverse to humans (e.g. underwater, space robotics and disaster response) or the complexity of the task demands human-level cognition and dexterity (e.g. robotic surgery and manufacturing). These systems typically use direct teleoperation at the motion level, and are usually limited to low-DOF arms and 2D perception. Improving dexterity and situational awareness demands new interaction and planning workflows. We explore the use of human-robot teaming through teleautonomy with assisted planning for remote control of a dual-arm dexterous robot for multi-step manipulation tasks, and conduct a within-subjects experimental assessment (n=12 expert users) to compare it with other methods, resulting in the following four conditions: 
\textbf{(A)} Direct teleoperation with imitation controller + 2D perception,
\textbf{(B)} Condition A + 3D perception,
\textbf{(C)} Teleautonomy interface teleoperation + 2D\&3D perception,
\textbf{(D)} Condition C + assisted planning.
The results indicate that this approach (D) achieves task times comparable with direct teleoperation (A,B) while improving a number of other objective and subjective metrics, including re-grasps, collisions, and TLX workload metrics. When compared to a similar interface but removing the assisted planning (C), D reduces the task time and removes a significant interaction with the level of expertise of the operator, resulting in a performance equalizer across users. {\footnotesize \href{https://sites.google.com/view/teleautonomy/}{Supplementary materials and accompanying video at {\color{blue} https://sites.google.com/view/teleautonomy/ }}}
\end{abstract}

\section{Introduction}
Technologies that enable performing manipulation of objects by controlling a robot in a remote
environment have been a key enabler of robotics in real world applications where, in addition to
navigation and inspection, interacting with the environment to change its state is required.
Telemanipulation is particularly relevant in environments where adverse circumstances make the
presence of humans impossible or undesirable, or where the contributions of autonomous systems in
collaboration with humans have the potential to improve the overall performance due to accuracy of
execution in time and space.  Examples of these requirements are found in a wide variety of
applications, such as disaster response, explosive ordnance disposal (EOD)
\cite{carruth2017challengestacticalteams} \cite{murphy2004human}, space robotics (robotic arm
\textit{Canadarm} on the ISS), underwater robotics (underwater manipulation for inspection
\cite{ocean_one} and disaster mitigation), and medical applications (surgical robotics
\cite{nichols2016framework}, teleoperation with \textit{Da Vinci} robot). 
These challenging environments and tasks that require high performance tend to require operators
with domain expertise (knowledge particular to the application) and formal training (expertise in
the use of the interfaces, devices, and methods developed and tested for the domain), and typically
are subject to constrained resources, such as limited time, \textit{in-situ} deployment, and
irreparable damage in case of failure.

Robotic systems able to operate in field conditions for these applications are largely based on
joint-by-joint teleoperation using switches and joysticks for motion control and 2D camera views for
perception \cite{chen2007human}. While direct teleoperation enables rapid deployment, this approach
has circumscribed deployed systems to the use of robotic arms with a low number of degrees of
freedom (DoF), such as the Packbot robot \cite{packbot}. Increased dexterity based on a higher
number of DoF faces scalability problems in terms of what is physically possible to control from
this type of interface. The scalability of direct teleoperation is also limited by decreasing
performance and instability as time delays increase.
Superior levels of dexterity and situational awareness require finding techniques that scale well
with the increased workload associated with controlling a higher number of DoF and with managing the
information contained in richer perception feedback such as 3D representations of the environment.

\begin{figure}[t]
  \centering          
  \includegraphics[scale=0.4]{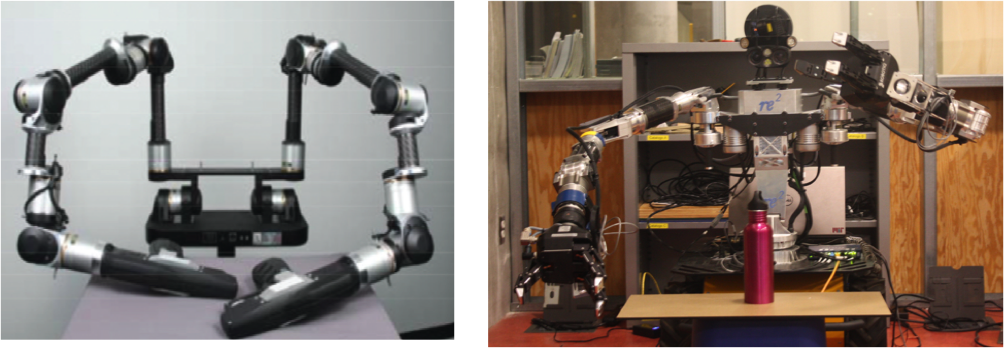}
  \caption[Photo of the Imitation Controller (IC) device]{Robot Hardware used in the study. The Imitation Controller (left) is a scaled
    kinematic replica of the robot (right). \vspace{-7mm}}
  \label{fig:ic_robot}
\end{figure}

The 2012-2015 DARPA Robotics Challenge (DRC) \cite{DRC_resultsandperspectives} served a large
testbed of multiple and competitive approaches to remote robot operation on simulated field
conditions. Multiple teams deployed fielded systems \cite{MITdrcTrials_JFR_2015_nobold}
\cite{RoboSimian_ROB:ROB21676} \cite{IHMCfinals_ROB:ROB21674} to conduct a remote robot through a
series of mobility and manipulation tasks inspired by challenges found during the response to the
Fukushima nuclear accident in 2011, such as turning a valve or opening a door 
\cite{fukushima}\cite{strickland2014fukushima}. These instances of telemanipulation
systems range in the autonomy spectrum from \textbf{teleoperation}, in which the human operator 
directly controls the movement of the remote robot or of a model of the robot \cite{whitney2020VRteleop}, to \textbf{teleautonomy}, 
in which the task is executed through an interaction workflow between the human 
operator and the robotic system
\cite{yanco_Trials}\cite{yanco_Finals}.

Previous work in the literature analyses the performance of the teams during the DRC Finals in terms
of interaction methods, robot characteristics, control methods, and sensor fusion. Results indicated
an increase in performance with increased human robot interaction patterns in terms of balancing
tasks between the operator and the robot \cite{yanco_Finals}. While the DRC competition was a
state-of-the-art demonstration of the multiple approaches to teleautonomy, and this detailed study
found advantages in using human robot teaming strategies, the competition conditions made it
difficult to conduct a controlled study.  In particular, teams had different numbers of operators,
each with a diverse set of roles within each operational framework. We sought to assess the task
performance in a controlled fashion with a \textit{single operator} managing all aspects of robot
operation.

Previous user studies in the field of surgical robotics have found teleoperation to produce the
fastest task times \cite{human_centered_debridement_GoldbergOkamura_IROS2015} but not the highest
accuracy results when compared to models of human-robot collaboration. The user study in
\cite{assesment_collamodels_okamura_icra2016} compared the use of teleoperation, supervised control,
traded control, and full autonomy in an inclusion segmentation task. Their results show faster task
completion times for teleoperation versus the other methods, while the metric of average force of
palpation over the body (less force is desired) was the highest in teleoperation. These previous
studies indicate a trade-off between the task execution speed and the accuracy as measured by
metrics relevant to the particular task.

The present study analyzes a similar space of human-robot collaboration models with various degrees
of autonomy, while focusing in the following operational constraints:  1) multiple sequential manipulation steps are required; 
2) steps are geometrically constrained; 
3) the perception feedback is limited to sensors on board the robot, as opposed to specialized sensor systems mounted
externally in a position tailored to the specific task; 
and 4) the tasks are performed by domain experts.
The \textbf{results} are consistent with previous findings in the literature 
of faster execution times for direct teleoperation (\textbf{A}) while exhibiting a similar trend in
the trade-off between task time and accuracy for methods that increase the level of robot assistance
and interaction (\textbf{B} and \textbf{C}) and experimentally find a human-robot collaboration
model that resulted in task times in the range of those produced by direct teleoperation while
significantly improving accuracy (\textbf{D}), overcoming the aforementioned trade-off for the first
time without expertly programmed sequences of motions. We also explore the interaction effects between the
subject's expertise type and the interface being used and find that while one teleautonomy method
exhibits a significant interaction,  causing differences in performance per expertise group
(\textbf{C}), increasing the level of autonomy on the robot’s side in a way that furthers the level
of collaboration (\textbf{D}) overcomes the performance limit correlated with the expertise level,
potentially offering a performance equalizer between domain experts with formal training in
different fields.

\section{Survey of Related Approaches}

In \textbf{direct teleoperation}, the robot moves simultaneously with the commanded motion using a
fixed mapping from master to slave. Multiple human-robot collaboration models that build on the base
of direct teleoperation have been proposed and evaluated \cite{SheridanBook}
\cite{Burstein96issuesin} \cite{tambe2002adjustablereal} \cite{sellner2006sliding} \cite{7281253}
\cite{assesment_collamodels_okamura_icra2016} \cite{Dragan_2012_formalizingassisteleop}. 
In particular, the method of \textit{shared control} aims to improve the usability of teleoperation
systems by continuously blending the input signal from the operator with a signal produced by an
autonomous system based on a prediction of the objective. This method has been
shown to increase task performance in a number of teleoperation settings
\cite{Dragan_2012_formalizingassisteleop} \cite{Shared_BCI_RSS_15} \cite{Shared_Hindsight_RSS_15}
\cite{AssistWithConstraintsShared_Dragan_2016}
\cite{jeon2020sharedlatent}, 
most commonly in applications where the  autonomous
assistance is meant to be complementary to operators with some deficit in operating the control
interface \cite{Shared_BCI_RSS_15}.

\textbf{Teleautonomy} approaches are motivated by the idea of a division of labor between the
operator and  the robotic system that maximizes their potentials,  such as the ability of the robot
to perform low-level perception and motion planning, and the operator's high-level task planning and
scene understanding. This division of labor is implemented through an interaction workflow that
enables information to flow between the two parts in order to make planning and execution decisions,
typically supervised by the human as the top-level decision maker. 
The \textit{teleautonomy} workflow is realized through a \textbf{teleautonomy interface}, which is a 
computer interface that affords on-line interactions with a human operator for operations related to
perception, planning, and control while immersed in a 3D world that represents the 
environment of the robot.
The most basic planning method on the teleautonomy interface is based on teleoperation, in which the
operator can specify the goal pose of the joints or of the end effectors of the robot and the system
computes the motion plan to achieve it. 

One approach to increase the level of autonomy in the teleautonomy interface is \textbf{assisted 
planning}, in which the robot has the ability to suggest motion plans to the operator automatically
in an on-line fashion without the operator indicating  the goal explicitly. Taking advantage of the
computation capabilities of the robot, it is possible for the system to autonomously compute motion
plans that accomplish the task, given that some information about the goal and objects involved can
be in the system \textit{a priori} or can be obtained from the operator through on-line queries. 
A first generation of assisted planning systems in this context deployed during the DRC was based on
template-based instances of pre-scripted motion plans executed with supervision from the operators
\cite{yanco_Finals}. 
For tasks known in advance, it is possible to program motions parametrized with
respect to objects in the scene
and use the human operator in the loop to
correctly instantiate these parameters in the scene
\cite{Director_JFR_Marion17_nobold}.

A higher level of integration of the human robot team involves overcoming the need for an expert
programmer to design sophisticated sequences of parametrized motions in advance. Learning from
demonstrations (LfD) is an approach designed to enable robots to learn how to execute manipulation
tasks from human demonstrations \cite{Survey_LfD_2009}. Integrating LfD into the system enables a
new concept of operations in which a skilled domain expert, not a programmer, can teach manipulation
skills to the robot and then execute these tasks remotely in a teleautonomy framework
\cite{darpino_clearn_ICRA17_nobold}. In previous work, multi-step manipulation tasks have been
learned from a single human demonstration using the algorithm C-LEARN
\cite{darpino_clearn_ICRA17_nobold}. This learning is done by leveraging accumulated knowledge about
how humans typically manipulate objects. Tasks are learned in terms of a sequence of steps and a set
of geometric constraints that define each step. After the learning phase, the learned task
representation is integrated with the teleautonomy framework for human-in-the-loop execution. This
strategy results in \textbf{teleautonomy with assisted planning}, in which at task execution time, 
the robot can plan for each learned step and produce
a motion suggestion for the human operator. The motions can be generated for new instances of the
same learned task where the geometry (position and orientation) of the objects is different from the
one in the demonstration.

This paper presents an evaluation of the performance of a human-robot teaming architecture based
on \textit{teleautonomy with assisted planning}, where the motion recommendations are generated using
C-LEARN  \cite{darpino_clearn_ICRA17_nobold}. 
This study compares four software/hardware architectures for control of a remote robot. The robot
used is MIT's \textit{Optimus} (Fig. \ref{fig:ic_robot}), a dual-arm manipulator (Highly Dexterous Manipulation System by
$re^2$) with two 3-fingers hands (by Robotiq) and a sensor suite with a Hokuyo sensor (Multisense SL
by Carnegie Robotics). The study uses 14-DoF in the arms (7-DoF per arm).

\section{Assisted Planning Approach}

\begin{figure}[t]
  \centering
  \includegraphics[scale=0.30]{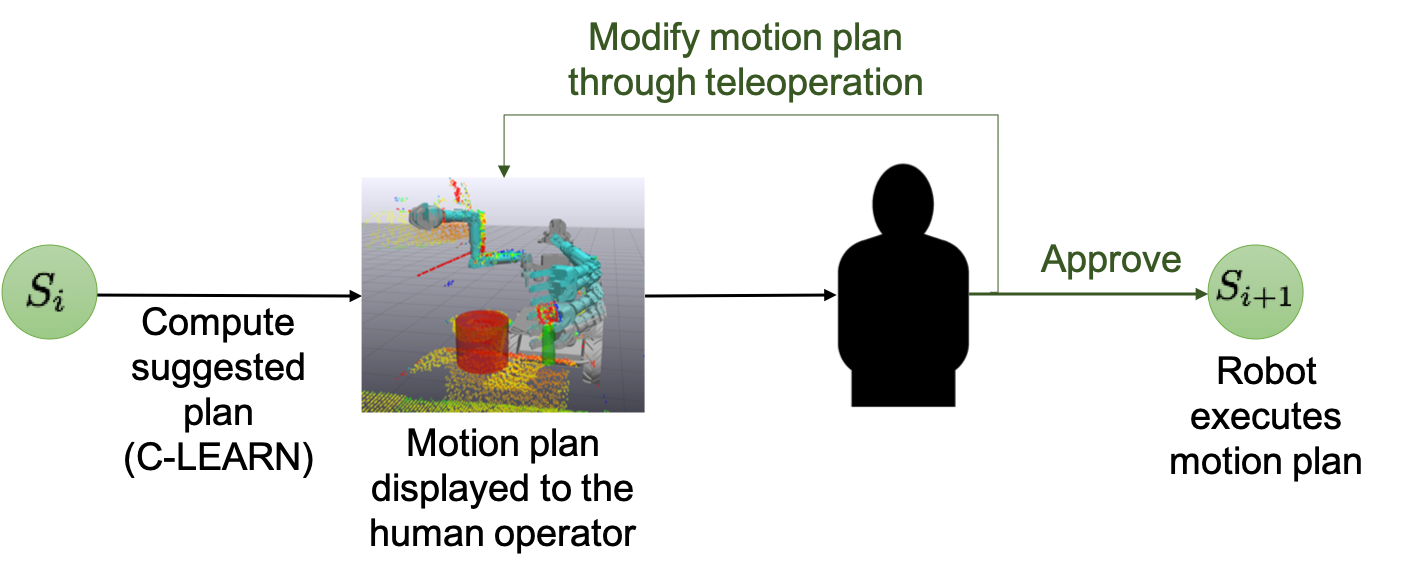}
  \caption{Human-Robot collaboration model for teleautonomy with task model learned through C-LEARN. \vspace{-3mm}}
  \label{fig:4_collab_model}
\end{figure}

The evaluated strategy for teleautonomy with assisted planning integrates a task representation learned through C-LEARN \cite{darpino_clearn_ICRA17_nobold} with a model of human robot collaboration for
telemanipulation. We take advantage of the keyframe-based representation used by C-LEARN, in which the task model consists of a series of geometric keyframes that provide the
information necessary for a planner to produce a motion plan given a new scenario. 
In this workflow (Fig.\ref{fig:4_collab_model}), the robot starts in the current keyframe and computes
a motion plan to reach the following keyframe in the current scene; this motion plan is displayed to
the human operator in the user interface. The operator has the ability to review this plan and approve it for 
robot execution in the remote environment.
Otherwise, the operator can reject the plan and perform
modifications by using end effector teleoperation in the interface. Note that this paper evaluates the performance of this interaction workflow that uses a learned task model, but not the capabilities of C-LEARN as a learning model.

\begin{figure}[!t]
  \centering
  \includegraphics[scale=0.3]{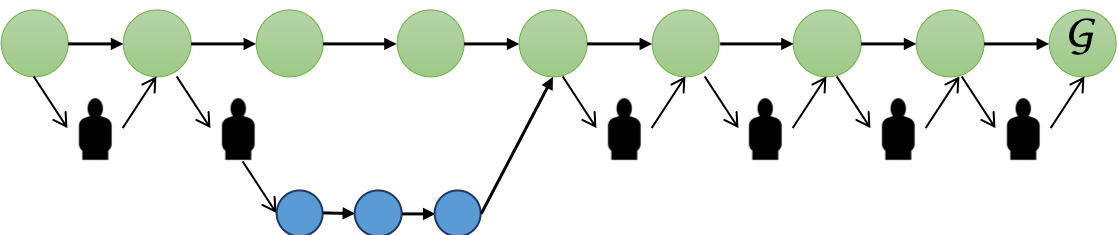}
  \caption[Workflow of planning and execution for a sequence of keyframes]{Workflow of planning and
    execution for a sequence of keyframes. The operator can accept a suggested motion plan (green keyframes) or
    perform teleoperation (blue keyframes) and return to suggested plans.\vspace{-5mm}}
  \label{fig:4_collab_model_flow}
\end{figure}

This workflow enables the operator to go from keyframe to keyframe as recommended by the robot
system, as illustrated by the green keyframes in Fig.\ref{fig:4_collab_model_flow}, or deviate from
the robot’s plan for a number of keyframes through teleoperation, as represented by the blue
keyframes. In the latter case, it is still possible to return to the sequence of suggestions from
the learned model, as long as the topology of the task still corresponds to the task learned
originally. 

\section{Conditions}

\begin{figure}[t]
  \centering          
  \includegraphics[width=0.45\textwidth]{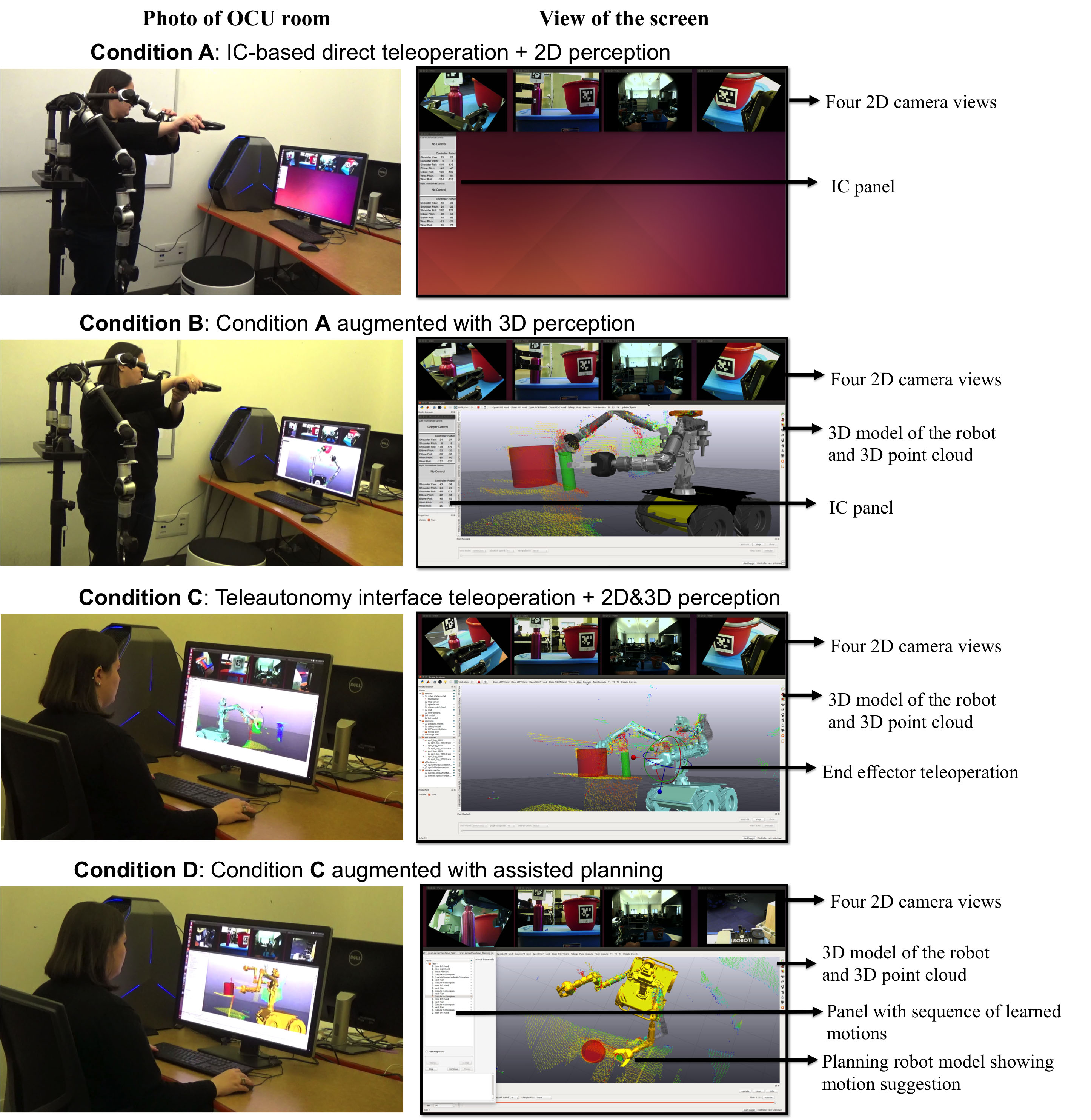}
  \caption[View of the OCU room and the content displayed on the computer monitor for each
  condition]{ {\small View of the OCU room (left column) and the content displayed on the
      computer monitor (right column) for each condition:  Condition \textbf{A}: IC-based direct
      teleoperation + 2D perception;  Condition \textbf{B}: Condition \textbf{A} augmented with 3D
      perception; Condition \textbf{C}: Teleautonomy interface teleoperation + 2D\&3D perception;
      Condition \textbf{D}: Condition \textbf{C} augmented with assisted planning. The operator had no direct line of sight with the robot in any condition. 
    } \vspace{-7mm}
  }
  \label{fig:conditions}
\end{figure}

The experimental assessment includes the following four conditions (Fig. \ref{fig:conditions}):
\textbf{(Condition A)} IC-based direct teleoperation + 2D perception,
\textbf{(Condition B)} Condition A augmented with 3D perception,
\textbf{(Condition C)} Teleautonomy interface teleoperation + 2D\&3D perception, and
\textbf{(Condition D)} Condition C augmented with assisted planning. The accompanying video shows the conditions and tasks used in this study \cite{PaperURLandVideo2020}.

Conditions \textbf{A} and \textbf{B} (Figure \ref{fig:conditions}(a)(b)) are based on \textbf{direct
  teleoperation} using an \textit{imitation controller (IC)} device  (Figure \ref{fig:ic_robot}). The
IC used as master is a passive device whose structure is a scaled kinematic replica of the robot
used as slave, which enables motion of the IC to be directly mapped to robot motion. 
We explore the use of the IC with two variants of perception: (\textbf{A}) 2D perception and
(\textbf{B}) 2D+3D perception, in which 3D perception consists of a view as described in
Figure \ref{fig:conditions}. 

\begin{figure}[t]
  \centering          
  \includegraphics[width=0.44\textwidth]{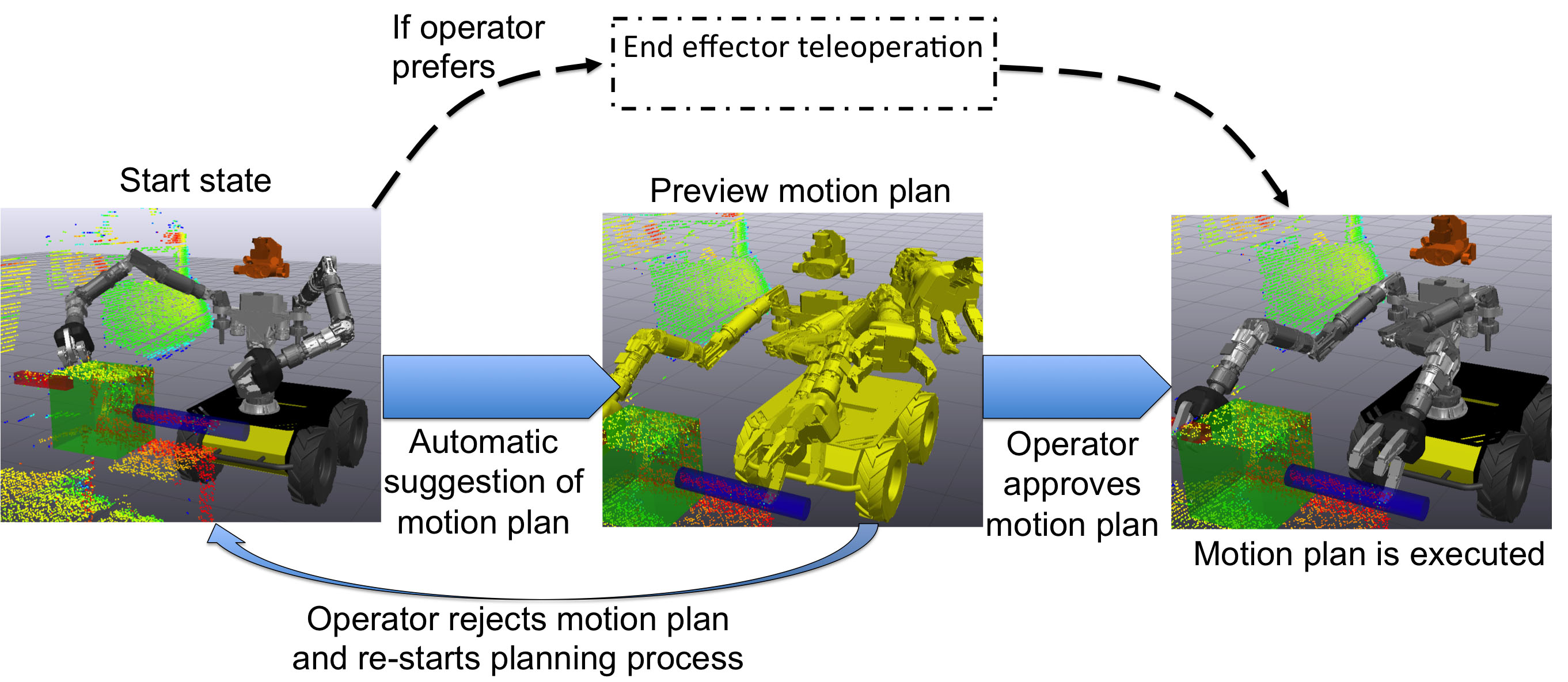}
  \caption{Workflow for assisted planning. Condition \textbf{D}. \vspace{-5mm}}
  \label{fig:AssistedPlanningFlow1}
\end{figure}

Conditions \textbf{C} and \textbf{D} are based on the \textbf{teleautonomy
  interface},  depicted in the screen views in Figure \ref{fig:conditions}(c)(d).
We use the interface \textit{Director} \cite{Director_JFR_Marion17_nobold}, the open-source user
interface developed by Team MIT to pilot the Atlas robot in the DARPA Robotics Challenge (DRC)
Finals. Through this interface, the operator has access to a 3D representation of the robot model
and the robot's environment (Figure \ref{fig:conditions}). This 3D representation is displayed in a 2D
monitor, similarly to the 3D environment used in video games or other robot interfaces such as RVIZ
from the Robot Operating System (ROS). In the teleautonomy interface, we experiment with the
following two planning workflows. Condition \textbf{C} is based in end-effector teleoperation, in
which the operator indicates the desired pose of the end effectors of the robot by either manually
dragging the virtual robot's hands or by positioning virtual floating hands within the 3D view in
the interface. Then the robot uses a motion planner to compute a motion plan to go from its current
configuration to a configuration that achieves the indicated poses of the end effectors. This motion
plan is previewed in the interface. Condition \textbf{D} uses assisted planning based on C-LEARN
\cite{darpino_clearn_ICRA17_nobold}\cite{CPDA_MIT_Thesis}, in which the system displays to the operator a series of
suggestions of motion plans automatically with the workflow illustrated in Fig. \ref{fig:AssistedPlanningFlow1}. Unlike IC-based teleoperation (conditions \textbf{A} and
\textbf{B}), where there is simultaneous motion of the master and the slave, in the teleautonomy
interface (conditions \textbf{C} and \textbf{D}), motion plans are first elaborated and previewed on
the interface and then sent to the robot for execution upon approval from the operator.  Note that
\textit{Director} is used for 3D perception on Conditions \textbf{B}, \textbf{C} and \textbf{D}.  In
\textbf{B}, Director affords only visualization, while in \textbf{C} and \textbf{D} it affords
visualization and interaction for robot control.
Both \textbf{C} and \textbf{D} use the same optimization-based motion planner \cite{MITdrcTrials_JFR_2015_nobold}
\cite{darpino_clearn_ICRA17_nobold}, available in Drake
\cite{DRAKE}, which uses the solver SNOPT \cite{SNOPT_2002}. The difference depends on the source of the goal specification
(operator vs. automatic).

\section{User Study Design and Protocol}

\begin{figure*}[t]
  \centering          
  \includegraphics[width=0.8\textwidth]{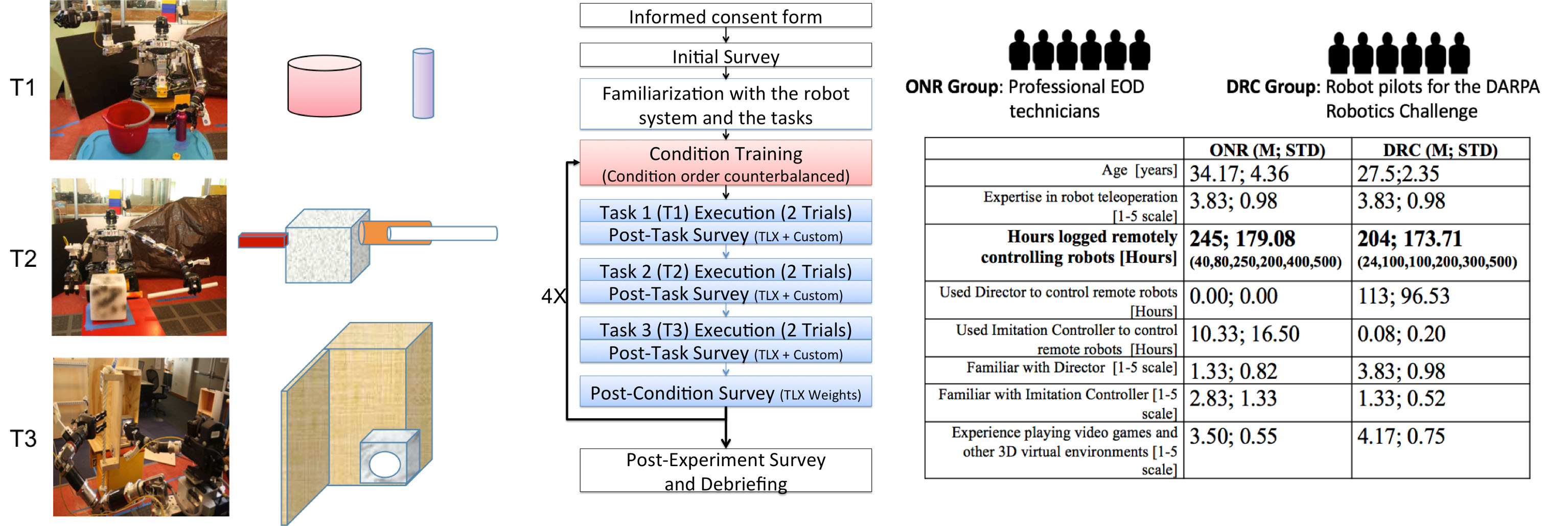}
  \caption{Manipulation tasks (left). Study protocol flow diagram (middle). Participants information as self-reported in initial survey (right). 
  \vspace{-5mm}}
  \label{fig:tasks_protocol_participants}
\end{figure*}

A within-subjects study with an expert population in robot telemanipulation was conducted to
evaluate the performance of four interfaces in three manipulation tasks.  
We collected objective
performance data during the task executions and subjective performance and satisfaction metrics with
a number of surveys. The study was conducted over the course of two days per participant. This time
permitted extensive training per condition and frequent breaks.  
All participants executed three \textit{Tasks} (T1, T2 and T3)
in all four \textit{Conditions} (A,B,C,D) during two successful \textit{Trials} of each task,
for a total of 24 runs per participant, and 288 runs in the complete user study.
Each trial used the same initial pose of the robot
and the same position
and orientation of the objects involved in each task.  
Each participant followed the protocol outlined in Figure \ref{fig:tasks_protocol_participants}(middle).
The study protocol and consent form were approved by the Institutional Review Board of the
Massachusetts Institute of Technology.  

\textbf{Participants and Assignment Method:}
The study was conducted with a total of 12 participants (11 males, 1 female, aged 24-41, M=30.83,
SD=4.82), recruited from a population with domain expertise and practical experience
with remote robot control. A summary of the self-reported expertise collected on the initial survey is presented in
Figure \ref{fig:tasks_protocol_participants}(right).
Their previous expertise is divided in two levels:
\textbf{ONR:} 6 participants (5 males, 1 female, aged 30-41, M=34.17, SD=4.36) with domain
  expert knowledge in the area of EOD. In particular, 3 of the participants in this group are
  professional EOD technicians. This group has extensive expertise in joint-by-joint teleoperation
  (primarily using switches and joysticks) of low-DOF robots (e.g., iRobot Packbot \cite{packbot},
  Foster-Miller Talon) using 2D perception only (from on-board cameras). This group is professionally
  trained to execute complex telemanipulation tasks under time pressure and safety concerns as
  required in EOD.
\textbf{DRC:} 6 participants (6 males, aged 24-30, M=27.5, SD=2.35) were remote robot
  operators during the DARPA Robotics Challenge (DRC, 2012-2015) for team MIT. This group has
  experience in remote control of high-DOF robots (e.g., Atlas robot by Boston Dynamics), with 3D
  perception through the user interface \textit{Director} \cite{Director_JFR_Marion17_nobold}. This
  group is trained to execute telemanipulation tasks under time pressure as required for the DRC
  competition.

The order in which participants of each expertise group (DRC and ONR) experienced the four
conditions was selected using a 4x4 Latin square. For each group, 4 participants completed one 4x4
Latin square, and the 2 remaining were assigned a random row. The Latin squares of each
expertise group were counter-balanced. A manipulation check showed no significance for the effect of
the order in which conditions were experienced over task time $(p=0.34)$.

\textbf{Tasks:} 

In each condition (randomized), three tasks (Fig. \ref{fig:tasks_protocol_participants}(left)) were executed in the same sequential order (not
randomized). We designed the tasks to be ordered in increasing order of difficulty according to
previous experimentation.
\textbf{Task 1 (T1):} Grasp the cylinder at the left, transport it, and release the cylinder
  inside a container. This task requires only one arm.
\textbf{Task 2 (T2):} Grasp the handle at the right side to secure the box, grasp the
  cylinder at the left, and extract the cylinder. This task requires dual arm manipulation with
  simultaneous contact with the same structure.
\textbf{Task 3 (T3):} Grasp the door handle (rope) at the left, open the cabinet, release the
  door, reach inside the cabinet at the right side, and push a button to turn on a light. This task
  requires dual arm manipulation and reaching into a confined space.

\textbf{Metrics and Statistical Analysis:}

\textbf{Objective Metrics:}
Total  Task  Time (from the moment the operator takes control of the interface to the moment the task is accomplished); 
Object of Interest (OOI) Moved (Number of times the robot moved the  OOI --manipulation  target  object-- outside  task  instructions); 
Collisions  with  OOI (Number  of  times the  robot  came  into  unintended  contact  with the OOI); 
Full Grasps Vs. Tip Grasps (Probability of issued grasp command to accomplish a full grasp or tip grasp);
Re-grasp (number of times the operator reattempted the  same  grasp). 
\textbf{Subjective Metrics:} 
NASA Task Load Index (TLX) \cite{NASA_TLX_1988}; 
HRI  Metrics  in  Post-Condition  Survey
  \cite{WorkingAlliance_horvath1989} \cite{hoffman_hri_metrics_2013HRIworkshop}
  \cite{HRImetricsJournal2019};  
Condition  Ranking  in  Post-Experiment  Survey. 
\textbf{Statistics:} 
For each
metric, a mixed effect model with randomized effects was fit to determine the effects of the \textit{Condition},
\textit{Expertise}, \textit{Task} and \textit{Trial} factors, as well as two-way interactions when appropriate\footnote{
Statistical support was provided by the Institute for Quantitative Social Science, 
Harvard University.}.
The following describes the general model
using the 
Wilkinson notation \cite{WilkinsonNotation1973}.
\begin{equation*}
\begin{aligned}
DV \text{{\raise.17ex\hbox{$\scriptstyle\sim$}}} &Trial + 
(Condition*Expertise) + (Condition*Task) + \\ 
&(Task*Expertise) + (1|Subject\_ID)
\end{aligned}
\end{equation*}
Due to space constraints, further details of the models, metrics, numerical results and significance are provided the supplementary materials \cite{PaperURLandVideo2020}.

\section{Analysis and Discussion of Results}

\begin{figure}[t]
  \centering          
  \includegraphics[scale=0.35]{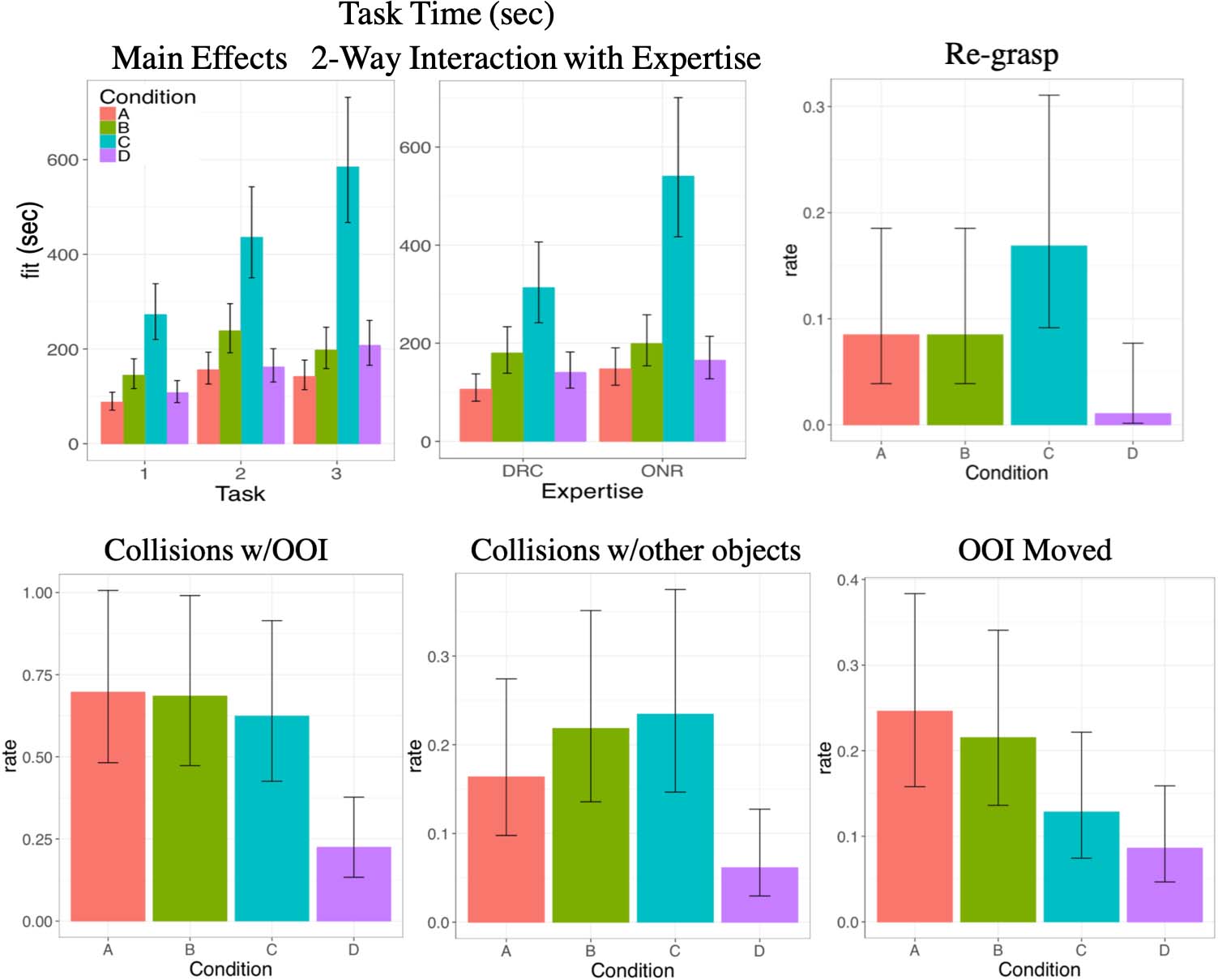}
  \caption[]
  {Main effects of Condition over objective metrics, and 2-Way interaction between Total Time and Expertise. Error bars show 95\% Confidence Intervals.
  \vspace{-5mm}} 
  \label{fig:objective_metrics}
\end{figure}

\textbf{Results:}
\textbf{Total Task Time:}
\textbf{(R1)} 
\condition{D} was significantly faster than \condition{C} for all task $(p<0.0001)$ and expertise
$(p<0.0001)$ groups.
\textbf{(R2)} 
\condition{D} total time is within a range comparable to teleoperation
\condition{A} and \condition{B} with no significant difference (Fig.\ref{fig:objective_metrics} (1,1)). The expected means follow the relation $A<D<B$ for all tasks and
expertise levels except for T3 where $D>B$ with no significant difference $(p=0.6316)$.
\textbf{(R3)}
 \condition{C} took more time than \condition{A},\condition{B}, and \condition{D} for all tasks and expertise groups $(p<0.0001)$.
\textbf{(R4)}
\condition{C} is the only that experimented a significant total time 2-way interaction between the
ONR and DRC expertise groups, taking and expected mean increase of 1.72 for the ONR group
$(p=0.0035)$ (Fig.\ref{fig:objective_metrics} (1,2)). 
\textbf{(R5)}
However, while using the same computer interface as
in \condition{C}, \condition{D} resulted in no significant time difference between the two expertise
groups $(p=0.3858)$. 
\textbf{(R6)}
For interface conditions (C,D), total time followed $T1<T2 (C,D: p<0.0001)$, and $T2<T3 (C:
p=0.0059; D: p=0.0191)$, while direct teleoperation conditions (A,B)  resulted in $T1<T3 (A:
p<0.0001; B: p=0.0044)$, and  $T3<T2 (A: p=0.3433; B: p=0.0665)$.
\textbf{(R7)}
Expected mean time of A is lower than B for all task $(T1,T2: p<0.0001; T3: p=0.0033)$ and expertise
levels $(DRC: p<0.0001, ONR: p=0.0015)$.
\textbf{(R8) OOI moved:}
The expected rate decreased as $A>B>C>D$ (Fig.\ref{fig:objective_metrics} (2,3)).
The rate for D is significantly lower than for both of the direct teleoperation $(A: p=0.0043; B: p=0.0188)$.
\textbf{(R9) Collision with OOI:}
The rate for D is
significantly lower than all conditions $(A: p=0.0478; B: p=0.0038; C: p=0.0021)$
\textbf{(R10) Grasping:}
The expected re-grasp rate difference between A and B was not detectable (p=1.000), whereas for interface
conditions D had a smaller rate than C $(p=0.0423)$
The expected probability of a full grasp (vs. tip grasp) follows the trend $D>C>B>A$, with
D significantly higher than A $(p=0.0371)$.
\textbf{(R11)}
The expected \textbf{TLX metrics score} of \condition{D} was significantly better than in conditions \condition{A},
\condition{B} and \condition{C} for both DRC and ONR expertise levels as follows:
\textbf{Mental Demand:} 
DRC $(A: p=0.0138;$ $B: p<0.0001;$ $C: p=0.0012)$,
ONR $(A: p=0.0002;$ $B: p=0.0004;$ $C: p<0.0001)$;
\textbf{Physical Demand:} 
DRC $(A: p<0.0001;$ $B: p<0.0001;$ $C: p=0.0011)$,
ONR $(A: p<0.0001;$ $B: p<0.0001;$ $C: p=0.0154)$;
\textbf{Temporal Demand:} 
DRC $(A: p=0.0565;$ $B: p=0.0165;$ $C: p=0.0121)$,
ONR $(A: p=0.0020;$ $B: p<0.0001;$ $C: p<0.0001)$;
\textbf{Effort:} 
DRC $(A: p<0.0001;$ $B: p<0.0001;$ $C: p<0.0001)$,
ONR $(A: p<0.0001;$ $B: p<0.0001;$ $C: p<0.0001)$;
\textbf{TLX Total Score:} 
DRC $(A: p<0.0001;$ $B: p=<0.0001;$ $C: p<0.0001)$,
ONR $(A: p=0.0307;$ $B: p=0.0001;$ $C: p<0.0001)$.

\textbf{Analysis:}

\textbf{The monotonic trade-off between task time and motion accuracy as human-robot co-activity
  increases can be reversed with assisted planning:}
The use of assisted planning in condition \condition{D} resulted in \textit{total task times}
comparable to both IC-based direct teleoperation conditions \condition{A} and \condition{B} for all
tasks and expertise levels (R2), whereas condition \condition{C} resulted in significantly longer task
times than all other conditions (R3). While \condition{D} times are comparable to \condition{A} and
\condition{B}, all other objective metrics were improved by \condition{D} 
(such as in R8, R9 and R10, showing an objective
improvement over the overall goal of the operator (to prioritize accuracy while minimizing task
time). This result has implications for the previous understanding of a monotonic trade-off between
execution speed and motion accuracy when moving from pure teleoperation to models of human-robot
collaboration that increase the level of co-activity according to the skills of each agent. In
particular, assisted planning in \textbf{D} does in fact achieve task times
comparable to teleoperation while improving accuracy (instead of improving accuracy at the cost of
slower task times). 

\textbf{Increased robot autonomy removes interaction effect of task time with expertise:}
Even though \condition{C} and \condition{D} use the same computer interface, total task time in
\condition{C} had a significant 2-way interaction of task time with expertise that \condition{D} did
not exhibit (R4 and R5), indicating that the use of assisted planning resulted in a performance equalizer 
between the two expertise groups in the use of the teleautonomy interface. Task time in IC-based
conditions \condition{A} and \condition{B} did not experience significant variations for expertise
groups, possibly due to the intuitiveness afforded by the IC.
This result has implications for the deployment of these system in real high-intensity domains,
where experts from different fields often come together to work on a given situation.

\textbf{Task difficulty interacts with interface type:}
Task difficulty for each task (T1, T2 and T3), as measured by task time, resulted in a different
ordering for IC-based conditions (\textbf{A} and \textbf{B}) than for teleautonomy conditions
(\textbf{C} and \textbf{D}). The results show that task time followed the relation $T1<T2<T3$ in
\condition{C} and \condition{D}, whereas it followed $T1<T3<T2$ in \condition{A} and \condition{B} (R6).
This interaction indicates that different manipulation maneuvers present different levels of
difficulty in the computer interface than in teleoperation with the IC.

\textbf{Total task time is heteroscedastic:}
The data on Total Task Time exhibited strong heteroscedasticity, which shows that the variance of
the total time increases as a function of increasing time. The finding of larger variability for
longer runs indicates diminishing performance predictability for longer runs, possibly due to
supervening circumstances in a particular task run or the dexterity of the operator.

\textbf{Task learning effect doesn't transfer across different interfaces:}
The main effect of Trial over task time was significant ($Trial2<Trial1$ $p<0.0001$), showing a learning effect for a given task
on a given condition. However, users did not become faster at a certain task as they repeated it in
different conditions throughout the course of the experiment (manipulation check $(p=0.34)$).

\textbf{The addition of 3D perception in teleoperation resulted in longer task times, while the
  expected higher accuracy was not significant:}
The addition of 3D perception in IC-based direct teleoperation conditions resulted in a total time
increase in \textbf{B} with respect to \condition{A} (R7). 
The expected benefits of the 3D perception on the improvement of the other metrics related to
manipulation accuracy did not result in significant changes in \condition{A} vs. \condition{B}. TLX
scores are favorable to \condition{A} but not significantly, possibly due to the workload associated
with the management of the 3D view. 
Note that while using the same physical device is used in \condition{A} and \condition{B}, the TLX
physical demand has a higher expected score for \condition{B}, possibly due to the increased time
per task. 
Furthermore, operators reported in the section for open comments that the addition of the 3D view in
teleoperation was very informative.
However, this addition did not result in observable benefits. We believe this result was hindered by
the complexity of adjusting the 3D view while operating the IC.

\textbf{TLX metrics show a decrease in workload for assisted planning:}
Subjective metrics for task workload (TLX) and human-robot collaboration are favorable to
\condition{D} in all categories (R11).  Computer interfaces show lower physical demand than IC-based
teleoperation conditions.\vspace{-2mm}

\section{Conclusions}

This paper presents an experimental assessment of a human-robot teaming model 
based on assisted planning 
for multi-step remote manipulation 
that leverages learned task models \cite{darpino_clearn_ICRA17_nobold} to compute and suggest motion plans to the operator.
We compare this system with three established models: direct teleoperation with 2D and 3D perception and
teleoperation based on a user interface. 
The study replicated real field conditions as much as
possible; all aspects of the system were implemented end-to-end (perception, planning, controls,
communications, user interfaces), and no Wizard of Oz technique was used. The study was conducted
with an expert population in control of mobile manipulators. The following are three main results of the study:
(1) The proposed method achieved task times comparable with direct teleoperation through an imitation controller, and improved task time significantly over using the same interface without the assisted planning component. 
(2) The proposed method significantly improved a number of objective (e.g. grasp quality,
  collisions, regraps) and subjective metrics (NASA TLX \cite{NASA_TLX_1988} and HRI metrics
  \cite{WorkingAlliance_horvath1989} \cite{hoffman_hri_metrics_2013HRIworkshop}
  \cite{HRImetricsJournal2019}). 
(3) The use of end effector teleoperation through the 3D user interface had a significant interaction with the previous expertise of the users. The addition of assisted planning (motion suggestions from the robot), while using the same interface, removed this interaction, resulting in a performance equalizer across users.

\bibliographystyle{ieeetr}

\input{suplementary.tex}

\end{document}

%% file: suplementary.tex
\newpage

\section{Supplementary Materials}

\section{Details of Experimental Conditions}

The four experimental conditions are detailed in this section. 
During motion execution in
all methods, the low-level controller of the robot used position control in joint angles space. The
robot arms were connected to constant power in all runs to avoid the possibility of  performance differences due to
decreasing battery levels during the course of the study.

\subsection{(Condition A) IC-based direct teleoperation + 2D perception:}
During \textit{IC-based direct teleoperation}, the operator manipulates the Imitation Controller
device (IC) (Figure \ref{fig:ic_robot}).  Joint angles from the IC are mapped directly to joint
angles in the real robot, which executes the commanded motion simultaneously with the IC. 
\textbf{2D Perception} provides the following four 2D camera live views: one camera mounted on each
wrist of the robot, one on the base, and one on the head.
Joint angles are streamed from the IC to the robot at 10 Hz, and camera feed is streamed from the
robot to the OCU screen at 15 frames per second (fps). A view of the interface is shown in Figure
\ref{fig:conditions}(a). Examples of the interface workflow, robot task execution and OCU room are shown in the accompanying video for all conditions. 
This condition provides a field benchmark, as it uses similar perception as in the systems fielded
today for real operations. In terms of the input device for teleoperation, field operations use
switches and joystick to control individual joints, but this methodology is used with three or four degrees of freedom (DoF). We
acknowledge that this method does not scale well with number of DoF and for this reason use instead
an imitation controller (IC) which enables control of all DoF or simply moving the end effector of the IC and leaving all the IC joints accommodate accordingly.

\subsection{(Condition B) Condition A augmented with 3D perception:}
This condition consists of the same IC-based teleoperation implementation as in condition
\textbf{A}, with the addition of 3D perception.
\textbf{3D perception} provides a live view of a 3D point cloud being sensed by a rotating Hokuyo
mounted on the head of the robot. 
An example is shown in Figure \ref{fig:conditions}(b).

In addition to the live 2D views available in Condition \textbf{A}, 
robot joint angles and the 3D point clouds are
streamed from the robot to the OCU for visualization.

\subsection{(Condition C)  Teleautonomy interface teleoperation + 2D\&3D perception} 
The teleautonomy interface enables the operator to command the robot through 
 \textbf{end effector teleoperation}.
Unlike IC-based direct teleoperation, where the robot moves simultaneously with the motion of the IC
input device, teleautonomy has a two-step workflow of planning an execution. Motion plans are first
composed on the interface and are executed only after approval. 

\begin{figure*}[h]
  \centering          
  \includegraphics[width=0.8\textwidth]{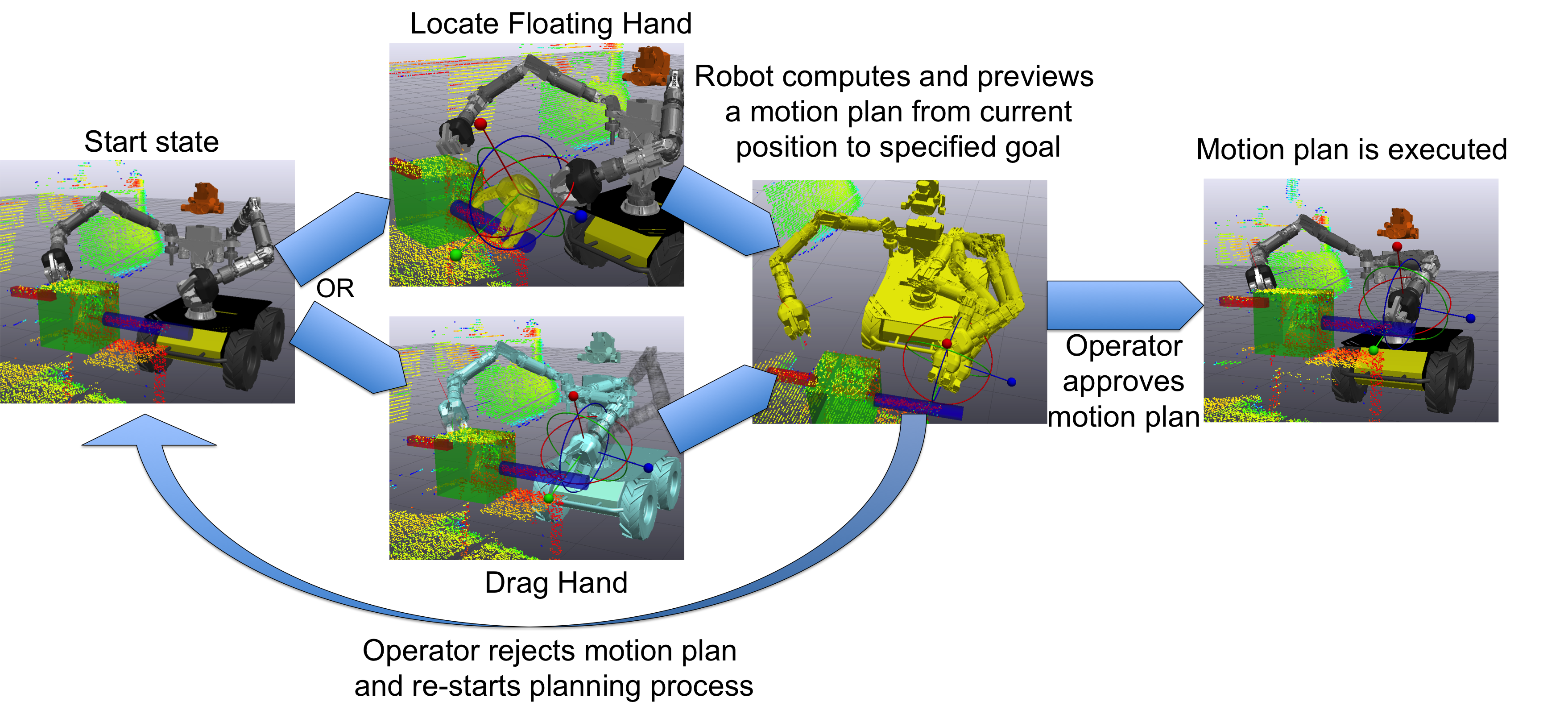}
  \caption[Workflow for end effector teleoperation in the teleautonomy interface]{Workflow for end
    effector teleoperation in the teleautonomy interface. Available in conditions  \textbf{C} and
    \textbf{D}.}
  \label{fig:EEteleopFlow}
\end{figure*}

This workflow for this condition is shown in Figure \ref{fig:EEteleopFlow}. The current state of the
robot and the environment is represented on the 3D view.  The operator specifies a desired goal
position and orientation of the end effectors of the robot. The goal can be specified by either
dragging the hands of the robot on the interface or by locating a floating hand. After the goal is
specified, the system automatically computes a motion plan for moving from its current configuration
to a configuration that satisfies the requested end effector pose. The system produces a 3D
animation of the computed motion plan, which is shown overlapped with the 3D representation of the
environment. The animation can be re-played by the operator as many times as needed. The operator
approves or rejects the motion plan. If approved, the motion plan is sent to the robot for
execution.

\subsection{(Condition D) Condition C augmented with assisted planning:} 
This condition focuses on \textbf{assisted planning}, in which the robot automatically suggests  a sequence
of motion plans to the operator. The workflow is illustrated in Figure
\ref{fig:AssistedPlanningFlow1}. Given a start state, the operator clicks a button to obtain the next
motion suggestion. The suggestion is previewed in animation, in the same fashion as in condition \textbf{C}. If the
operator approves, the motion plan is sent to the robot for execution. The operator is free to recur
to the end effector teleoperation workflow (\textbf{C}) at any moment. 

In this study, the motion recommendations are generated from a pre-learned model using C-LEARN
\cite{darpino_clearn_ICRA17_nobold}. For the sake of uniformity, the pre-learned model used is the
same for all participants. Participants had not seen in advance the motions that the robot would
suggest in each task. 

In \textbf{C}, the operator uses the visualization of the sensor data (point cloud and virtual
objects) as a reference to specify the desired position of the end effectors. In assisted planning
in \textbf{D}, the robot computes automatically the goal of the end effectors with respect to frames
in the objects, which are obtained from the same perception subsystem. The underlying information
about the position of objects is the same in both conditions.

\section{Training} 

Participants were trained for each condition immediately before the test time in that condition. No
time limit was assigned for training sessions to enable participants to achieve the expected level
of proficiency before proceeding with the test tasks. All participants had a guided training
session, during which the same instructor provided the same sequence of technical instructions
interleaved with practice time for each concept or technique involved. After the guided training
session, participants practiced using all capabilities in the interface to execute a training task.
Participants were allowed to proceed to the test session after being able to execute the training
task successfully without technical errors in the use of the interface and having expressed feeling
comfortable with it. The training task consisted of grasping, transporting, and releasing a cylinder
over a table. The task was practiced with both arms, and the objects were located so that motions
had to take place in a region of the workspace similar to the one used in the test tasks. During
training, participants had available live video feedback of the robot room from cameras external to
the robot with the purpose of facilitating training.

\section{Evaluation Metrics}

\subsection{Objective Metrics}

\subsubsection{Total Task Time} 
Total task time spans the time from the moment the operator takes control of the interface to the
moment the task is accomplished.
\subsubsection{Object of Interest Moved}  
Number of times the robot moved the object of interest (manipulation
target object) outside task instructions. For example, due to undesired displacements during pushing
or lifting.
\subsubsection{Collisions with Object of Interest} 
Number of times the robot came into unintended contact with the object of interest.
\subsubsection{Full Grasps Vs. Tip Grasps} 
Every grasp was classified into two categories.  Full grasps are defined as stable grasps where the
object is in contact with the palm of the hand and all three fingers achieved closure. A tip grasp
is defined as an issued grasp command that failled to satisfy the full grasps conditions.
\subsubsection{Re-grasps} 
Number of times the operator reattempted the same grasp. For example, the operator commanded the
hand to close with the intention of grasping an object, but the hand closed in 
free space due to incorrect positioning of the end effector and, as a consequence, had to re-attempt
the grasp.

\subsection{Subjective Metrics}
\subsubsection{NASA Task Load Index (TLX)} 
Mental Demand, Physical Demand, Temporal Demand, Effort and TLX Total Score as defined by the NASA \textbf{T}ask \textbf{L}oad Inde\textbf{X} \cite{NASA_TLX_1988}. 
\subsubsection{HRI Metrics in Post-Condition Survey} 
\begin{figure}[h!]
  \centering          
  \includegraphics[scale=0.9]{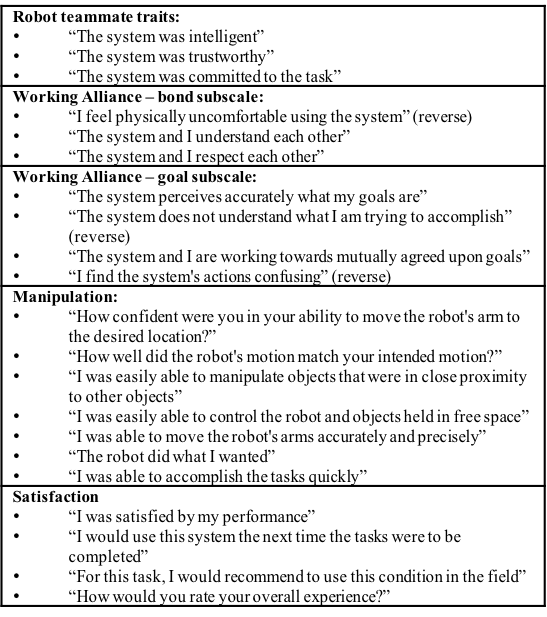}
  \caption{HRI questions in post-condition assessment}
  \label{table:Subjective_HRI_questions}
\end{figure}

A number of subjective metrics related to the fluency and performance of human-robot collaboration
were collected through a between-tasks questionnaire (1-7 Likert scale).
Table \ref{table:Subjective_HRI_questions} shows the questions related to aspects of human-robot
collaboration (HRC) and perception of performance, following the guidance of metrics for 
human-robot collaboration in \cite{WorkingAlliance_horvath1989}
\cite{hoffman_hri_metrics_2013HRIworkshop}, as well as custom metrics.

\subsubsection{Condition Ranking in Post-Experiment Survey} %
\begin{figure}[h!]
  \centering          
  \includegraphics[scale=0.9]{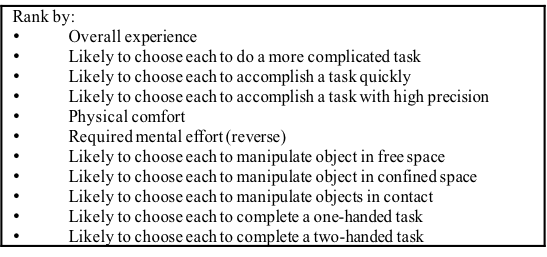}
  \caption{Rankings post-experiment assessment}
  \label{table:Subjective_rankings_questions}
\end{figure}

After completing the study, participants were asked to rank all conditions according to their
preferences regarding nine performance-related aspects and one overall final ranking.
Questions are itemized in Table \ref{table:Subjective_rankings_questions}.

\section{Statistical Models}

\subsection{Independent variables (IV)}
\textbf{\textit{Condition}} (interfaces \condition{A}, \condition{B}, \condition{C} and
\condition{D} as categorical variable);
\textbf{\textit{Task}} (Tasks T1, T2 and T3 as categorical variable);
\textbf{\textit{Expertise}} (Expertise levels DRC and ONR as categorical variable);
\textbf{\textit{Trial}} (Trial number Trial 1 and Trial 2 as categorical variable);
\textbf{\textit{SubjectID}} (Subject identifier 1 to 12 as categorical variable).

\subsection{Dependent variables (DV)}
Objective metrics (\textit{Total Task Time}, \textit{OOI
  moved}, \textit{Collision with OOI}, \textit{Collision with other objects}, \textit{Full Grasps Vs.
  Tip Grasps}, \textit{Regraps}) and subjective metrics (seven \textit{TLX scores}, seven \textit{HRI
  scores}, ten \textit{final rankings}).

\subsection{Models}
The effect of the IVs considered in this study on each DV is
determined by fitting a \textbf{mixed effect model} individually for each DV. 
Each model included random
effects of the factor \textbf{SubjectID} with a random intercept $(1|SubjectID)$, to account for
different responses per subject according to their baseline level. 
Each model tested fixed effects
of \textit{Condition}, \textit{Task}, \textit{Expertise}, \textit{Trial} for main effects, simple
effects, and 2-way interactions as detailed below. 
Holm adjustment was applied to the models.
The following describes each model and equation
using the Wilkinson notation \cite{WilkinsonNotation1973}.

For \textbf{Total Task Time}, the data exhibited strong heteroscedasticity. To account for it, we
used a generalized linear mixed effect model with a gamma distribution (logarithmic link).

\vspace{-5mm}
\begin{equation*}
\begin{aligned}
DV \text{{\raise.17ex\hbox{$\scriptstyle\sim$}}} &Trial + 
(Condition*Expertise) + (Condition*Task) + \\ 
&(Task*Expertise) + (1|Subject\_ID)
\end{aligned}
\end{equation*}

For all other objective metrics, we hypothesized the main effects of Trial, Task, Condition and
Expertise. Two-way interactions were not hypothesized and tested not significant. The model for
these metrics followed: 
\begin{equation*}
\begin{aligned}
DV \text{{\raise.17ex\hbox{$\scriptstyle\sim$}}} Trial + 
Condition + Expertise + Task + 
(1|Subject\_ID)
\end{aligned}
\end{equation*}
except for \textit{Re-grasps} where the factor Task was removed due to non-convergence. The metrics
\textbf{Object of interest moved},  \textbf{Collision with OOI}, \textbf{OOI moved}, and
\textbf{Re-graps} were fitted individually using a generalized linear mixed effect model with a
Poisson distribution. 
The metric \textbf{Full Grasp} was modeled with a generalized linear effect model with a binomial
(logarithmic link), which enables use of the counts of full and tip grasps to model the probability
of a full grasp.

The \textbf{NASA Task Load Index (TLX)} includes seven scores in a scale 0-100. We fit a generalized
linear mixed model with a binomial distribution.
We hypothesize the existence of two-way interactions for the factors Condition, Expertise and Task.
Note that Trial is not a factor because the TLX questionnaire was administered only after
participants finished the second trial of each task. The model results in the following equation:
\begin{equation*}
\begin{aligned}
DV \text{{\raise.17ex\hbox{$\scriptstyle\sim$}}} &
(Condition*Expertise) + (Condition*Task) + \\ 
&(Task*Expertise) + (1|Subject\_ID)
\end{aligned}
\end{equation*}

\subsubsection{Post-Condition Survey}  
Questions are grouped in five categories: Robot teammate traits
\cite{hoffman_hri_metrics_2013HRIworkshop}, Working Alliance – bond subscale
\cite{WorkingAlliance_horvath1989} \cite{hoffman_hri_metrics_2013HRIworkshop},  Working Alliance –
goal subscale \cite{WorkingAlliance_horvath1989} \cite{hoffman_hri_metrics_2013HRIworkshop},
Manipulation, Perception, Satisfaction. 
Answers were in 1-7 Likert Scale.
We report the Cronbach's $\alpha$ measure of consistency for
each group and fit a linear mixed model to each group with the following equation.
\begin{equation*}
\begin{aligned}
DV \text{{\raise.17ex\hbox{$\scriptstyle\sim$}}} &
(Condition*Expertise) + (Condition*Task) + \\ 
&(Task*Expertise) + (1|Subject\_ID)
\end{aligned}
\end{equation*}
\subsubsection{Post-Experiment Survey} 
An ordinal cumulative link mixed model was fit to each ranking question in the post-experiment
assessment. We
hypothesized a 2-way interaction of Condition and Expertise. 
Note that Task and Trial are not
factors for this model because this information was queried by the end of the user study. The model
uses the following equation:
\begin{equation*}
\begin{aligned}
DV \text{{\raise.17ex\hbox{$\scriptstyle\sim$}}} &
(Condition*Expertise) + (1|Subject\_ID)
\end{aligned}
\end{equation*}

Data analysis was performed using R (R version 3.4.2) using custom code and R packages 
lsmeans\_2.25     
ordinal\_2015.6-28 
ggplot2\_2.2.1     
effects\_4.0-0    
optimx\_2013.8.7   
robustlmm\_2.1-3   
nlme\_3.1-131      
lattice\_0.20-35   
lmerTest\_2.0-33  
car\_2.1-5         
lme4\_1.1-14.
No treatment for outliers was performed.

\section{Detailed Results}

We investigate and compare the performance of conditions \condition{A}, \condition{B}, \condition{C}
and \condition{D} through a number of objective and subjective metrics defined below. For each
metric, a mixed effect model was fit to determine the effects of the \textit{Condition},
\textit{Expertise}, \textit{Task} and \textit{Trial} factors, as well as two-way interactions when
pertinent.

\subsection{\textbf{Objective Metrics}}

For total task time, we hypothesized the following effects: main effect of $Trial$ to account for
the learning effect of using the same interface in the same task during the two trials; 2-way
interaction for $(Condition*Expertise)$ to account for time performance differences in the same
condition between participants from the ONR and DRC groups; interaction for $(Condition*Task)$ to
account for a given condition enabling a level of dexterity in a specific manipulation skill present
in one task but not others that resulted in time performance differences;  $(Task*Expertise)$ to
account for possible previous expertise causing a time performance difference across different
tasks. Three-way interactions were not hypothesized, and no significance was found when tested. 

For all other objective metrics, we hypothesized the main effects of Trial for potential learning
effects, Task to account for different geometries changing the likelihood of different events,
Condition to account for the interface, and Expertise to account for performance differences due to
pre-existent skills.

\hfill \break
\subsubsection{\textbf{Total Task Time}} 
Total task time spans the time from the moment the operator takes control of the interface to the
moment the task is accomplished.  The results show significant two-way interactions for the pairs
Condition*Task $(Pr(>Chisq)=0.0046)$ and Condition*Expertise $(Pr(>Chisq)=0.0009)$  and a simple
effect of Trial $(Pr(>Chisq)<2.2e-16)$.

\textbf{Interface methods: How does C compare with D?}
\condition{D} was significantly faster than \condition{C} for all task $(p<0.0001)$ and expertise
$(p<0.0001)$ groups.

\textbf{Direct teleoperation methods: How does A compare with B?}
Expected mean time of A is lower than B for all task $(T1,T2: p<0.0001; T3: p=0.0033)$ and expertise
levels $(DRC: p<0.0001, ONR: p=0.0015)$.

\textbf{How do interface conditions (C,D) compare 
  with teleoperation conditions (A,B)?}

\condition{C} took more time than \condition{A},\condition{B}, and \condition{D} for all tasks and
expertise groups $(p<0.0001)$.

The Total Task Time of \condition{D} is within a range comparable to the teleoperation conditions
\condition{A} and \condition{B}. The expected means follow the relation $A<D<B$ for all tasks and
expertise levels except for T3 where $D>B$ with no significant difference $(p=0.6316)$. 

\textbf{How does Condition interact with Expertise?}
\condition{C} is the only condition that experimented a significant total time variation between the
ONR and DRC expertise groups, taking and expected mean increase of 1.72 for the ONR group
$(p=0.0035)$. This performance difference might be attributable to the previous experience in
controlling robots through a computer interface. However, while using the same computer interface as
in \condition{C}, \condition{D} resulted in no significant time difference between the two expertise
groups $(p=0.3858)$, providing evidence that assisted planning makes complex interfaces more
accessible to different training backgrounds. Similarly, for direct teleoperation conditions
\condition{A} and \condition{B}, expertise resulted in no significant differences in performance
$(A: p=0.0745; B: p=0.5887)$.

\textbf{How does Condition interact with Task?}
For interface conditions (C,D), total time followed $T1<T2 (C,D: p<0.0001)$, and $T2<T3 (C:
p=0.0059; D: p=0.0191)$, while direct teleoperation conditions (A,B)  resulted in $T1<T3 (A:
p<0.0001; B: p=0.0044)$, and  $T3<T2 (A: p=0.3433; B: p=0.0665)$.

\textbf{How does Expertise interact with Task?}
No significant interaction was found.

\textbf{Is there a learning effect between trials of the same task?}
The second trial resulted in faster times than the first trial when averaged across all expertise
and task levels $p<0.0001$. We attribute this result to a learning effect on how to execute a
particular task using a particular interface.  However, a manipulation check with respect to total
task time showed that this learning effect was not significant across conditions $(p=0.34)$, meaning
the learning effect of task $X$ with interface $Y$ did not carry over to task $X$ with interface
$Z$.

\hfill \break
\subsubsection{\textbf{Object of Interest Moved}}  
This metric is defined as the number of times the robot moved the object of interest (manipulation
target object) outside task instructions. For example, due to undesired displacements during pushing
or lifting. \textbf{Main effect of Condition:} Significant main effect found
$(Pr(>Chisq)=0.002034)$. The expected rate of moving the object of interest decreased as $A>B>C>D$.
The rate for interface condition D is significantly lower than for both of the direct teleoperation
conditions $(A: p=0.0043; B: p=0.0188)$. \textbf{Main effect of Task:} Significant main effect found
$(Pr(>Chisq)<2.2e-16)$. Task 2 had a significant increase compared with tasks 1 and 2 $(p<0.0001)$.
\textbf{Main effect of Expertise:} Significant main effect found $(Pr(>Chisq)=0.008907)$. The DRC
group had lower expected rates than the ONR group $(p=0.0089)$. \textbf{Main effect of Trial:} No
significant main effect found $(Pr(>Chisq)=0.563903)$.

\hfill \break
\subsubsection{\textbf{Collisions with Object of Interest}} 
Number of times the robot came into unintended contact with the object of interest.

\textbf{Main effect of Condition:} 
Significant main effect found $(Pr(>Chisq)=9.776e-05)$.  The rate for interface condition D is
significantly lower than for both of the direct teleoperation conditions and the other interface
condition C $(A,B: p=0.0001; C: p=0.0006)$.

\textbf{Main effect of Task:} Significant main effect found $(Pr(>Chisq)=7.246e-05)$. Task 2 had a
significant increase compared with Task 1 and Task 3 $(p=0.0008)$.

\textbf{Main effect of Expertise:} No significant main effect found $(Pr(>Chisq)=0.30306)$. The
expected rate follows $DRC<ONR$. 

textbf{Main effect of Trial:} No significant main effect found $(Pr(>Chisq)=0.08888)$. The expected
rate follows $Trial 2 < Trial 1.$

\textbf{Collisions with Other Objects} %
Number of times the robot came into unintended contact with the objects in the environment other
than the target object. For example, collisions with the table.

\textbf{Main effect of Condition:} 
Significant main effect found $(Pr(>Chisq)=0.002727)$.  The rate for interface condition D is
significantly lower than both of the direct teleoperation conditions and the other interface
condition C $(A: p=0.0478; B: p=0.0038; C: p=0.0021)$. \textbf{Main effect of Task:} Significant
main effect found $(Pr(>Chisq)=7.246e-05)$. Task 3 had a significant increase compared with Tasks 1
and 2 $(p<0.0001)$. \textbf{Main effect of Expertise:} No significant main effect found
$(Pr(>Chisq)=0.524511)$. The expected rate follows $DRC<ONR$. \textbf{Main effect of Trial:} No
significant main effect found $(Pr(>Chisq)=0.191910)$. The expected rate follows $Trial 2 < Trial
1.$

\hfill \break
\subsubsection{\textbf{Full Grasps Vs. Tip Grasps}} %
Every grasp was classified into two categories.  Full grasps are defined as stable grasps where the
object is in contact with the palm of the hand and all three fingers achieved closure. A tip grasp
is defined as a grasp that fails to satisfy the full grasps conditions.

\textbf{Main effect of Condition:} Main effect found trending towards significance
$(Pr(>Chisq)=0.050630)$.  The expected probability of a good grasp follows the trend $D>C>B>A$, with
D significantly higher than direct teleoperation condition A $(p=0.0371)$.

\textbf{Main effect of Task:} Significant main effect found $(Pr(>Chisq)=0.002212)$. Task 3 had a
good grasp probability bigger than task 1 and task 2 $(T1: p=0.0018; T2: p=0.0083)$.

\textbf{Main effect of Expertise:} No significant main effect found $(Pr(>Chisq)=0.107252)$. The
expected probability follows $DRC>ONR$. \textbf{Main effect of Trial:} No significant main effect
found $(Pr(>Chisq)=0.208448)$. The expected probability follows $Trial 2 > Trial 1.$

\hfill \break
\subsubsection{\textbf{Re-grasps}} %
Number of times the operator reattempted the same grasp. For example, the operator commanded the
hand to close with the intention of grasping an object, but the hand closed in 
free space due to incorrect positioning of the end effector and, as a consequence, had to re-attempt
the grasp. 

\textbf{Main effect of Condition} Significant main effect found $(Pr(>Chisq)=0.024559)$.  The
expected rate difference between A and B was not detectable (p=1.000), whereas for interface
conditions D had a smaller rate than C $(p=0.0423)$. \textbf{Main effect of Task:} No convergence
for this factor. 

\textbf{Main effect of Expertise:} Significant main effect found $(Pr(>Chisq)=0.006104)$. The
expected rate follows $DRC<ONR (p=0.0061)$.

\textbf{Main effect of Trial:} No significant main effect found $(Pr(>Chisq)=0.477918)$. The
expected rate follows $Trial 2 < Trial 1.$

\hfill \break
\hfill \break
\hfill \break
\hfill \break
\hfill \break
\hfill \break
\subsection{\textbf{Subjective Metrics}}

\begin{figure*}[h]
  \centering          
  \includegraphics[scale=0.35]{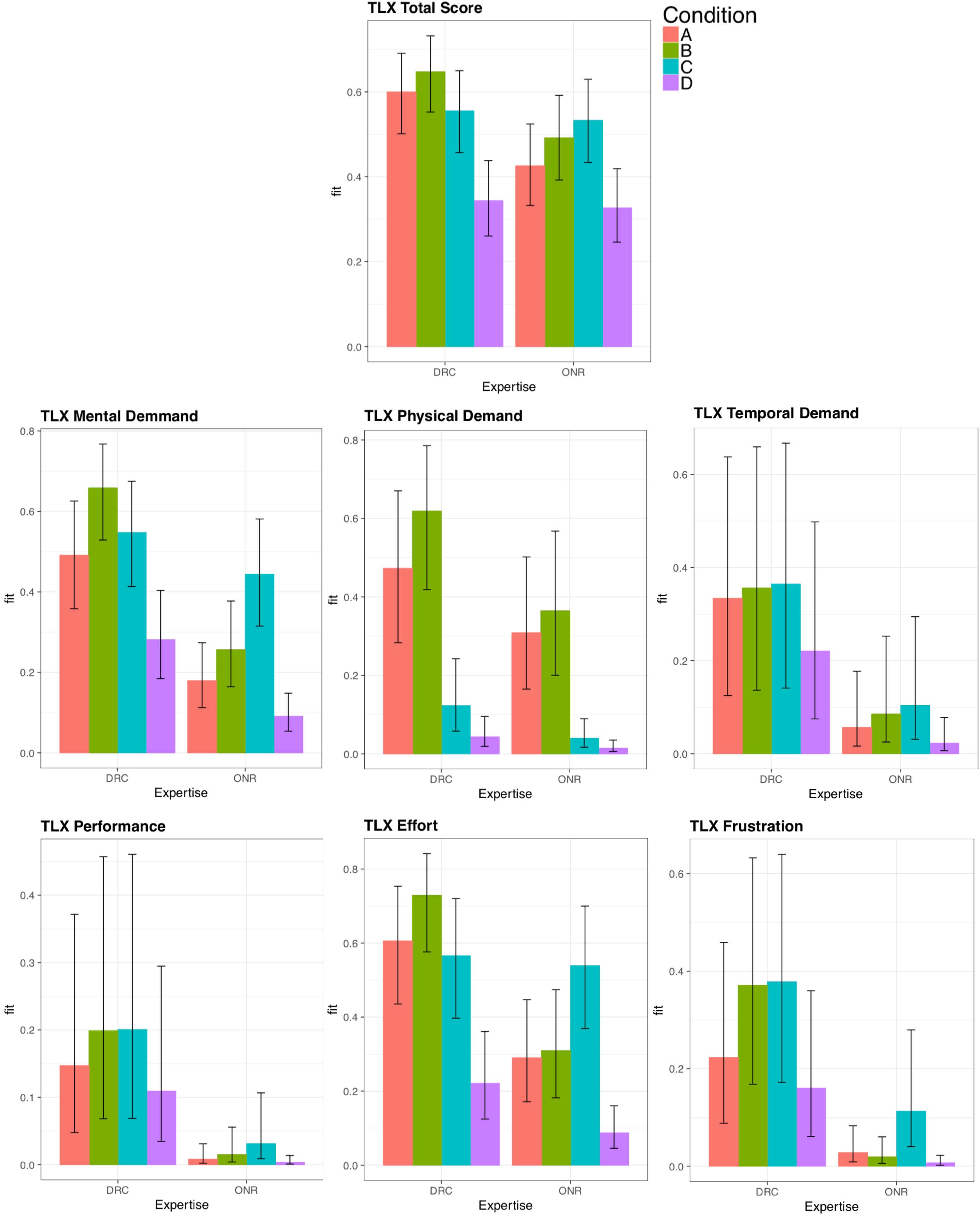}
  \caption{Results of TLX evaluation. Error bars show 95\% Confidence Intervals.}
  \label{fig:Subjective_TLX}
\end{figure*}

\subsubsection{\textbf{NASA Task Load Index (TLX)}}  
Results from the NASA TLX evaluation \cite{NASA_TLX_1988} are shown in Figure
\ref{fig:Subjective_TLX}.

\textbf{How does D compare to all other conditions?}
The expected score of \condition{D} was significantly better than in conditions \condition{A},
\condition{B} and \condition{C} for both DRC and ONR expertise levels for the following TLX metrics.

\textbf{Mental Demand:} 
DRC $(A: p=0.0138; B: p<0.0001; C: p=0.0012)$,
ONR $(A: p=0.0002; B: p=0.0004; C: p<0.0001)$;
\textbf{Physical Demand:} 
DRC $(A: p<0.0001; B: p<0.0001; C: p=0.0011)$,
ONR $(A: p<0.0001; B: p<0.0001; C: p=0.0154)$;
\textbf{Temporal Demand:} 
DRC $(A: p=0.0565; B: p=0.0165; C: p=0.0121)$,
ONR $(A: p=0.0020; B: p<0.0001; C: p<0.0001)$;
\textbf{Effort:} 
DRC $(A: p<0.0001; B: p<0.0001; C: p<0.0001)$,
ONR $(A: p<0.0001; B: p<0.0001; C: p<0.0001)$;
\textbf{TLX Total Score:} 
DRC $(A: p<0.0001; B: p=<0.0001; C: p<0.0001)$,
ONR $(A: p=0.0307; B: p=0.0001; C: p<0.0001)$.
The expected score for the TLX performance metric in condition \condition{D} was also lower than in
\condition{A} and \condition{B} for both expertise levels with no significance and lower than
\condition{C} for the ONR group (p=0.0001) and the DRC group with no significance. For the TLX
frustration metric, \condition{D} had a better expected score than \condition{A}, \condition{B}, and
\condition{C}, with significance in four of the contrasts: DRC $(B: p=0.0479; C: p=0.0479)$; ONR
$(A: p=0.0226; and C: p<0.0001)$.

\textbf{Physical demand of IC-based direct teleoperation:}
\condition{A} and \condition{B} resulted in significantly higher physical demand when compared to
both computer-based interfaces (\condition{C} and \condition{D}) for both expertise groups
$(p<0.0001)$.

\textbf{Interaction between Condition and Expertise}  
A number of TLX metrics had a significant 2-way interaction.
\textbf{TLX Total Score} $(Pr(>Chisq)=0.024559)$:
Both IC-based teleoperation conditions (\condition{A} and \condition{B}) had significant differences
between the DRC and ONR groups (A: p=0.144; B: p=0.0267) with a better score in the ONR group,
whereas the computer interface conditions (\condition{C} and {D}) did not result in significant
differences per expertise group.  The expected \textbf{total score} follows the relation $D<C<A<B$
for the DRC group, and $D<A<B<C$ for the ONR group (a lower score is better). \textbf{Frustration}
level had significantly better scores for the ONR group for all conditions  $(A: p=0.0043; B:
p<0.0001; C: p=0.0475; D: p=0.0001)$, as well as \textbf{temporal demand} $(A: p=0.0203; B:
p=0.0524; C: p=0.0801; D: p=0.0066)$, and \textbf{performance} $(A: p=0.0011; B: p=0.0027; C:
p=0.0243; D: p=0.0002)$.

\cleardoublepage

\onecolumn

\begin{multicols}{2}

\subsubsection{\textbf{Post-Condition Survey}} 

\begin{figure}[h!]
  \centering          
  \includegraphics[scale=0.35]{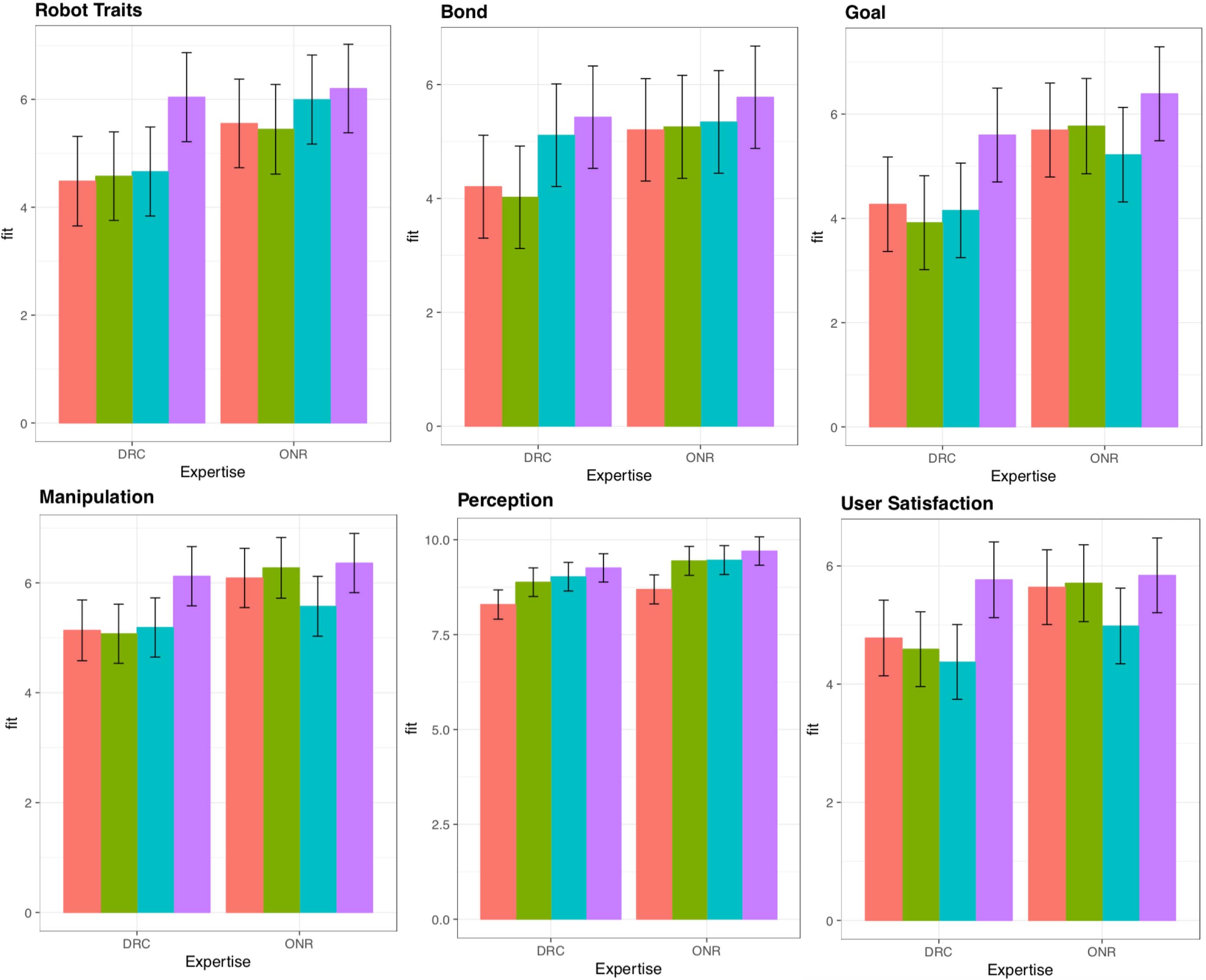}
  \caption{Post-condition survey subjective metrics. 1-7 Scale. Error bars show 95\% Confidence Intervals.}
  \label{fig:subjective_conditionHRI}
\end{figure}

A number of subjective metrics related to the fluency and performance of human-robot collaboration
were collected through a between-tasks questionnaire (1-7 Likert scale). Results are shown in Figure
\ref{fig:subjective_conditionHRI}  and summarized as follows.

\textbf{Robot teammate traits}: 
Cronbach's $\alpha = 0.7219$. Participants rated the robot team traits in \condition{D}  better than
in \condition{A}, \condition{B}  and \condition{C} $(p<0.0001)$ in the DRC group and better than
\condition{A} and \condition{B}  $(A: p=0.0353; B: p=0.0149)$ in the ONR group, within which these
traits also rated better than \condition{C} not significantly $(p=0.8010)$. 
\textbf{Working Alliance – Bond subscale}:
Cronbach's $\alpha = 0.5547$. Participants rated this metric in \condition{D} better than in
\condition{A} and \condition{B} in the DRC group $(p<0.0001)$ and the ONR group $(A: p=0.0068; B:
p=0.0611)$.

\textbf{Working Alliance – Goal subscale}:
Cronbach's $\alpha = 0.8452$. \condition{D} rated better than \condition{A}, \condition{B} and
\condition{C} in the DRC group (p<0.0001) and better than C for the ONR group $(p=0.0005)$.

\textbf{Manipulation}:
Cronbach's $\alpha = 0.9014$.
\condition{D} scored better than \condition{A}, \condition{B} and \condition{C} in the DRC group
$(A: p=0.0003; B: p=0.0001; C: p=0.0004)$. \condition{D} and \condition{B} scored better than
\condition{C} for the ONR group $(B: p=0.0259; C: p=0.0063)$.

\textbf{Perception}:
Cronbach's $\alpha = 0.4007$. In items related to perception, Condition \condition{A} (2D
perception) was rated significantly lower than \condition{B} (2D\&3D perception) $(DRC: p=0.0009;
ONR: p<0.0001)$, \condition{C} and \condition{D} $(p<0.0001)$. This is the only metric in the
post-condition assessment where there is detectable difference between IC-based conditions
\condition{A} and \condition{B} in either expertise group.

\textbf{Satisfaction}:  
Cronbach's $\alpha = 0.9319$. The DRC group reported higher satisfaction with \condition{D} than
with \condition{A}, \condition{B} and \condition{C} $(A: p=0.0034; B: p=0.0003, C: p<0.0001)$,
whereas ONR did compared to \condition{C} $(p=0.0187)$. The expected mean score of condition
\condition{D} is better than in all other conditions and expertise levels in all the metrics of the
post-condition assessment.

\end{multicols}

\cleardoublepage

\begin{multicols}{2}
\subsubsection{\textbf{Post-Experiment Survey}} 
\begin{figure}[h]
  \centering          
  \includegraphics[scale=0.74]{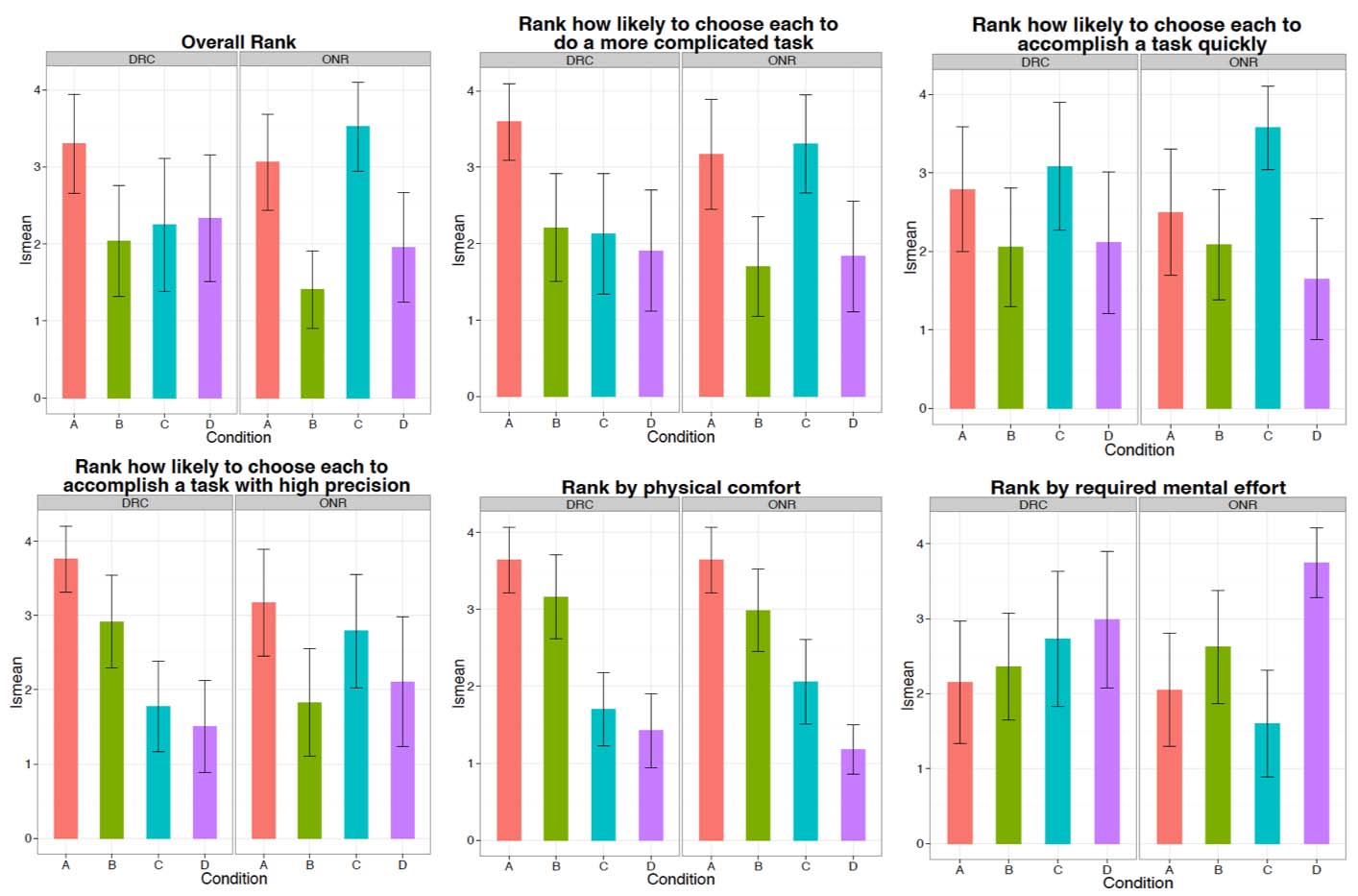}
  \caption{Post-Experiment Rankings Survey. Error bars show 95\% Confidence Intervals.}
  \label{fig:Subjective_FinalSurveyRankings}
\end{figure}

After completing the study, participants were asked to rank all conditions according to their
preferences regarding nine performance-related aspects and one overall final ranking. Ranking
results are presented in Figure \ref{fig:Subjective_FinalSurveyRankings}. The Cronbach's $\alpha$
consistency measure among all requested rankings is 0.8694. In the ONR group, the \textbf{Overall
  final ranking} showed no significant difference between \condition{B} and \condition{D}
$(p=0.4283)$, both of which ranked better than \condition{A} $(B: p=0.0003; D: p=0.0638)$ and
\condition{C} $(B: p<0.0001; D: p=0.0035)$. In the DRC group, \condition{B}, \condition{C} and
\condition{D} had no significant difference, ranking better than \condition{A}.
\end{multicols}

%% file: main.bbl
\begin{thebibliography}{10}

\bibitem{carruth2017challengestacticalteams}
D.~W. Carruth and C.~L. Bethel, ``Challenges with the integration of robotics
  into tactical team operations,'' in {\em Applied Machine Intelligence and
  Informatics (SAMI), 2017 IEEE 15th International Symposium on},
  pp.~000027--000032, IEEE, 2017.

\bibitem{murphy2004human}
R.~R. Murphy, ``Human-robot interaction in rescue robotics,'' {\em IEEE
  Transactions on Systems, Man, and Cybernetics, Part C (Applications and
  Reviews)}, vol.~34, no.~2, pp.~138--153, 2004.

\bibitem{ocean_one}
O.~Khatib, X.~Yeh, G.~Brantner, B.~Soe, B.~Kim, S.~Ganguly, H.~Stuart, S.~Wang,
  M.~Cutkosky, A.~Edsinger, P.~Mullins, M.~Barham, C.~R. Voolstra, K.~N.
  Salama, M.~L'Hour, and V.~Creuze, ``Ocean one: A robotic avatar for oceanic
  discovery,'' vol.~23, pp.~20--29, 2016.

\bibitem{nichols2016framework}
K.~A. Nichols and A.~M. Okamura, ``A framework for multilateral manipulation in
  surgical tasks,'' {\em IEEE Transactions on Automation Science and
  Engineering}, vol.~13, no.~1, pp.~68--77, 2016.

\bibitem{chen2007human}
J.~Y. Chen, E.~C. Haas, and M.~J. Barnes, ``Human performance issues and user
  interface design for teleoperated robots,'' {\em IEEE Transactions on
  Systems, Man, and Cybernetics, Part C (Applications and Reviews)}, vol.~37,
  no.~6, pp.~1231--1245, 2007.

\bibitem{packbot}
B.~M. Yamauchi, ``{PackBot}: a versatile platform for military robotics,'' in
  {\em Proc. SPIE 5422, Unmanned Ground Vehicle Technology}, vol.~5422,
  pp.~228--237, 2004.

\bibitem{DRC_resultsandperspectives}
E.~Krotkov, D.~Hackett, L.~Jackel, M.~Perschbacher, J.~Pippine, J.~Strauss,
  G.~Pratt, and C.~Orlowski, ``The darpa robotics challenge finals: Results and
  perspectives,'' {\em Journal of Field Robotics}, vol.~34, no.~2,
  pp.~229--240, 2017.

\bibitem{MITdrcTrials_JFR_2015_nobold}
M.~Fallon, S.~Kuindersma, S.~Karumanchi, M.~Antone, T.~Schneider, H.~Dai,
  C.~P\'{e}rez-D'Arpino, R.~Deits, M.~DiCicco, D.~Fourie, T.~Koolen, P.~Marion,
  M.~Posa, A.~Valenzuela, K.-T. Yu, J.~Shah, K.~Iagnemma, R.~Tedrake, and
  S.~Teller, ``An architecture for online affordance-based perception and
  whole-body planning,'' {\em Journal of Field Robotics}, vol.~32, no.~2,
  pp.~229--254, 2015.

\bibitem{RoboSimian_ROB:ROB21676}
S.~Karumanchi, K.~Edelberg, I.~Baldwin, J.~Nash, J.~Reid, C.~Bergh, J.~Leichty,
  K.~Carpenter, M.~Shekels, M.~Gildner, D.~Newill-Smith, J.~Carlton,
  J.~Koehler, T.~Dobreva, M.~Frost, P.~Hebert, J.~Borders, J.~Ma, B.~Douillard,
  P.~Backes, B.~Kennedy, B.~Satzinger, C.~Lau, K.~Byl, K.~Shankar, and
  J.~Burdick, ``Team robosimian: Semi-autonomous mobile manipulation at the
  2015 darpa robotics challenge finals,'' {\em Journal of Field Robotics},
  vol.~34, no.~2, pp.~305--332, 2017.

\bibitem{IHMCfinals_ROB:ROB21674}
M.~Johnson, B.~Shrewsbury, S.~Bertrand, D.~Calvert, T.~Wu, D.~Duran,
  D.~Stephen, N.~Mertins, J.~Carff, W.~Rifenburgh, J.~Smith,
  C.~Schmidt-Wetekam, D.~Faconti, A.~Graber-Tilton, N.~Eyssette, T.~Meier,
  I.~Kalkov, T.~Craig, N.~Payton, S.~McCrory, G.~Wiedebach, B.~Layton,
  P.~Neuhaus, and J.~Pratt, ``Team {IHMC}'s lessons learned from the {DARPA}
  robotics challenge: Finding data in the rubble,'' {\em Journal of Field
  Robotics}, vol.~34, no.~2, pp.~241--261, 2017.

\bibitem{fukushima}
K.~Nagatani, S.~Kiribayashi, Y.~Okada, K.~Otake, K.~Yoshida, S.~Tadokoro,
  T.~Nishimura, T.~Yoshida, E.~Koyanagi, M.~Fukushima, and S.~Kawatsuma,
  ``Emergency response to the nuclear accident at the fukushima daiichi nuclear
  power plants using mobile rescue robots,'' {\em Journal of Field Robotics},
  vol.~30, no.~1, pp.~44--63, 2013.

\bibitem{strickland2014fukushima}
E.~Strickland, ``Fukushima's next 40 years,'' {\em IEEE Spectrum}, vol.~51,
  no.~3, pp.~46--53, 2014.

\bibitem{whitney2020VRteleop}
D.~Whitney, E.~Rosen, E.~Phillips, G.~Konidaris, and S.~Tellex, ``Comparing
  robot grasping teleoperation across desktop and virtual reality with ros
  reality,'' in {\em Robotics Research}, pp.~335--350, Springer, 2020.

\bibitem{yanco_Trials}
H.~A. Yanco, A.~Norton, W.~Ober, D.~Shane, A.~Skinner, and J.~Vice, ``Analysis
  of human-robot interaction at the {DARPA} robotics challenge trials,'' {\em
  Journal of Field Robotics}, vol.~32, no.~3, pp.~420--444, 2015.

\bibitem{yanco_Finals}
A.~Norton, W.~Ober, L.~Baraniecki, E.~McCann, J.~Scholtz, D.~Shane, A.~Skinner,
  R.~Watson, and H.~Yanco, ``Analysis of human--robot interaction at the
  {DARPA} robotics challenge finals,'' {\em The International Journal of
  Robotics Research}, pp.~483--513, 2017.

\bibitem{human_centered_debridement_GoldbergOkamura_IROS2015}
K.~A. Nichols, A.~Murali, S.~Sen, K.~Goldberg, and A.~M. Okamura, ``Models of
  human-centered automation in a debridement task,'' in {\em Intelligent Robots
  and Systems (IROS), 2015 IEEE/RSJ International Conference on},
  pp.~5784--5789, IEEE, 2015.

\bibitem{assesment_collamodels_okamura_icra2016}
K.~E. Kaplan, K.~A. Nichols, and A.~M. Okamura, ``Toward human-robot
  collaboration in surgery: performance assessment of human and robotic agents
  in an inclusion segmentation task,'' in {\em Robotics and Automation (ICRA),
  2016 IEEE International Conference on}, pp.~723--729, IEEE, 2016.

\bibitem{SheridanBook}
T.~B. Sheridan, {\em Telerobotics, Automation, and Human Supervisory Control}.
\newblock Cambridge, MA, USA: MIT Press, 1992.

\bibitem{Burstein96issuesin}
M.~H. Burstein, B.~Beranek, N.~Inc, and D.~V. Mcdermott, ``Issues in the
  development of human-computer mixed-initiative planning,'' in {\em Cognitive
  Technology}, pp.~285--303, Elsevier, 1996.

\bibitem{tambe2002adjustablereal}
M.~Tambe, P.~Scerri, and D.~V. Pynadath, ``Adjustable autonomy for the real
  world,'' {\em Journal of Artificial Intelligence Research}, vol.~17, no.~1,
  pp.~171--228, 2002.

\bibitem{sellner2006sliding}
B.~Sellner, F.~W. Heger, L.~M. Hiatt, R.~Simmons, and S.~Singh, ``Coordinated
  multiagent teams and sliding autonomy for large-scale assembly,'' {\em
  Proceedings of the IEEE}, vol.~94, no.~7, pp.~1425--1444, 2006.

\bibitem{7281253}
S.~Jain, A.~Farshchiansadegh, A.~Broad, F.~Abdollahi, F.~Mussa-Ivaldi, and
  B.~Argall, ``Assistive robotic manipulation through shared autonomy and a
  body-machine interface,'' in {\em Rehabilitation Robotics (ICORR), 2015 IEEE
  International Conference on}, pp.~526--531, 2015.

\bibitem{Dragan_2012_formalizingassisteleop}
A.~Dragan and S.~Srinivasa, ``Formalizing assistive teleoperation,'' in {\em
  Robotics: Science and Systems}, 2012.

\bibitem{Shared_BCI_RSS_15}
K.~Muelling, A.~Venkatraman, J.-S. Valois, J.~Downey, J.~Weiss, S.~Javdani,
  M.~Hebert, A.~Schwartz, J.~Collinger, and A.~Bagnell, ``Autonomy infused
  teleoperation with application to bci manipulation,'' 2015.

\bibitem{Shared_Hindsight_RSS_15}
S.~Javdani, S.~Srinivasa, and J.~A.~D. Bagnell, ``Shared autonomy via hindsight
  optimization,'' in {\em Proceedings of Robotics: Science and Systems}, (Rome,
  Italy), 2015.

\bibitem{AssistWithConstraintsShared_Dragan_2016}
N.~Mehr, R.~Horowitz, and A.~Dragan, ``Inferring and assisting with constraints
  in shared autonomy,'' in {\em Conference on Decision and Control (CDC)},
  2016.

\bibitem{jeon2020sharedlatent}
H.~J. Jeon, D.~Losey, and D.~Sadigh, ``Shared autonomy with learned latent
  actions,'' in {\em Proceedings of Robotics: Science and Systems (RSS)}, 2020.

\bibitem{Director_JFR_Marion17_nobold}
P.~Marion, M.~Fallon, R.~Deits, A.~Valenzuela, C.~P\'{e}rez-D'Arpino, G.~Izatt,
  L.~Manuelli, M.~Antone, H.~Dai, T.~Koolen, J.~Carter, S.~Kuindersma, and
  R.~Tedrake, ``Director: A user interface designed for robot operation with
  shared autonomy,'' {\em Journal of Field Robotics}, vol.~34, no.~2,
  pp.~262--280, 2017.

\bibitem{Survey_LfD_2009}
B.~D. Argall, S.~Chernova, M.~Veloso, and B.~Browning, ``A survey of robot
  learning from demonstration,'' {\em Robotics and Autonomous Systems},
  vol.~57, no.~5, pp.~469 -- 483, 2009.

\bibitem{darpino_clearn_ICRA17_nobold}
C.~P\'{e}rez-D'Arpino and J.~A. Shah, ``{C-LEARN}: Learning geometric
  constraints from demonstrations for multi-step manipulation in shared
  autonomy,'' in {\em IEEE ICRA 2017}, 2017.

\bibitem{PaperURLandVideo2020}
{\em Project Website and accompanying video}.
\newblock 2020.
\newblock \url{https://sites.google.com/view/teleautonomy/}.

\bibitem{CPDA_MIT_Thesis}
C.~Pérez-D’Arpino, ``Hybrid learning for multi-step manipulation in
  collaborative robotics,'' {\em Massachusetts Institute of Technology Ph.D.
  Thesis}, 2019.
\newblock \url{https://dspace.mit.edu/handle/1721.1/122740}.

\bibitem{DRAKE}
R.~Tedrake, ``Drake: A planning, control, and analysis toolbox for nonlinear
  dynamical systems,'' 2014.
\newblock \url{http://drake.mit.edu}.

\bibitem{SNOPT_2002}
P.~E. Gill, W.~Murray, and M.~A. Saunders, ``{SNOPT}: An {SQP} algorithm for
  large-scale constrained optimization,'' {\em SIAM journal on optimization},
  vol.~12, no.~4, pp.~979--1006, 2002.

\bibitem{NASA_TLX_1988}
S.~G. Hart and L.~E. Staveland, ``Development of {NASA-TLX} (task load index):
  Results of empirical and theoretical research,'' {\em Advances in
  psychology}, vol.~52, p.~139–183, 1988.

\bibitem{WorkingAlliance_horvath1989}
A.~O. Horvath and L.~S. Greenberg, ``Development and validation of the working
  alliance inventory.,'' {\em Journal of counseling psychology}, vol.~36,
  no.~2, p.~223, 1989.

\bibitem{hoffman_hri_metrics_2013HRIworkshop}
G.~Hoffman, ``Evaluating fluency in human-robot collaboration,'' in {\em
  International conference on human-robot interaction (HRI), workshop on human
  robot collaboration}, vol.~381, pp.~1--8, 2013.

\bibitem{HRImetricsJournal2019}
G.~Hoffman, ``Evaluating fluency in human–robot collaboration,'' {\em IEEE
  Transactions on Human-Machine Systems}, vol.~49, no.~3, pp.~209--218, 2019.

\bibitem{WilkinsonNotation1973}
G.~N. Wilkinson and C.~E. Rogers, ``Symbolic description of factorial models
  for analysis of variance,'' {\em Applied Statistics}, vol.~22, no.~3,
  pp.~392--399, 1973.

\end{thebibliography}
